\documentclass[10pt,journal,compsoc]{IEEEtran}

\usepackage{times}
\usepackage{times}
\usepackage{epsfig}
\usepackage{graphicx}
\usepackage{amsmath}
\usepackage{amssymb}
\usepackage{bm}
\usepackage{overpic}
\usepackage{multirow}
\usepackage{booktabs}       % professional-quality tables
\usepackage{array}
\usepackage{makecell}
\usepackage{paralist}
\usepackage{amsthm}
\usepackage{mathrsfs}
\newcommand{\PreserveBackslash}[1]{\let\temp=\\#1\let\\=\temp}
\newcolumntype{C}[1]{>{\PreserveBackslash\centering}p{#1}}
\newcolumntype{R}[1]{>{\PreserveBackslash\raggedleft}p{#1}}
\newcolumntype{L}[1]{>{\PreserveBackslash\raggedright}p{#1}}

\usepackage[colorlinks,
linkcolor=blue,
anchorcolor=blue,
citecolor=blue,
]{hyperref}
\usepackage{url}

\ifCLASSOPTIONcompsoc
\usepackage[nocompress]{cite}
\else
\usepackage{cite}
\fi

\def\etal{\emph{et al}.~}
\def\ie{i.e.,~} % that is, in other words
\def\eg{\emph{e.g.},~}

\def\fst#1{\textbf{#1}} %in Table
\def\rev#1{#1} %in Table

\ifCLASSINFOpdf
\else
\fi

\begin{document}
	%
	% paper title
	% Titles are generally capitalized except for words such as a, an, and, as,
	% at, but, by, for, in, nor, of, on, or, the, to and up, which are usually
	% not capitalized unless they are the first or last word of the title.
	% Linebreaks \\ can be used within to get better formatting as desired.
	% Do not put math or special symbols in the title.
	\title{Instance-Level Relative Saliency Ranking with Graph Reasoning}

	\author{Nian~Liu,~\IEEEmembership{Member,~IEEE,}
		Long~Li,
		Wangbo~Zhao,
		Junwei~Han,~\IEEEmembership{Senior~Member,~IEEE,}
		and~Ling~Shao,~\IEEEmembership{Fellow,~IEEE}
		\IEEEcompsocitemizethanks{\IEEEcompsocthanksitem N. Liu and L. Shao are with the Inception Institute of Artificial Intelligence, Abu Dhabi, UAE. E-mail: liunian228@gmail.com, ling.shao@inceptioniai.org
			\IEEEcompsocthanksitem L. Li, W. Zhao, and J. Han are with School of Automation, Northwestern Polytechnical University, Xi'an, China, E-mail: \{longli.nwpu,wangbo.zhao96,junweihan2010\}@gmail.com
			\IEEEcompsocthanksitem Junwei Han is the corresponding author.
		}
	}

	% The paper headers
	\markboth{IEEE Transactions on Pattern Analysis and Machine Intelligence}%
	{Shell \MakeLowercase{\textit{et al.}}: Bare Demo of IEEEtran.cls for Computer Society Journals}

	\IEEEtitleabstractindextext{%
		\begin{abstract}
			Conventional salient object detection models cannot differentiate the importance of different salient objects. Recently, two works have \rev{been} proposed to detect saliency ranking by assigning different degrees of saliency to different objects. However, one of these models cannot differentiate object instances and the other focuses more on sequential attention shift order inference. In this paper, we investigate a practical problem setting that requires simultaneously segment salient instances and infer their relative saliency rank order. We present a novel unified model as the first end-to-end solution, where an improved Mask R-CNN is first used to segment salient instances and a saliency ranking branch is then added to infer the relative saliency. For relative saliency ranking, we build a new graph reasoning module by combining four graphs to incorporate the instance interaction relation, local contrast, global contrast, and a high-level semantic prior, respectively. A novel loss function is also proposed to effectively train the saliency ranking branch. Besides, a new dataset and an evaluation metric are proposed for this task, aiming at pushing forward this field of research. Finally, experimental results demonstrate that our proposed model is more effective than previous methods. We also show an example of its practical usage on adaptive image retargeting.
		\end{abstract}
		
		% Note that keywords are not normally used for peerreview papers.
		}

	% make the title area
	\maketitle

	\IEEEraisesectionheading{\section{Introduction}\label{sec:introduction}}
	
	\IEEEPARstart{S}{alient} object detection (SOD) aims 
	%has been widely explored for several years 
	to detect objects that attract people's attention in visual scenes. Various models have been proposed for this task. In recent years, deep learning-based models in particular, such as \cite{liu2016dhsnet,hou2017dss,liu2018picanet,zhang2018bmp,wang2019salient,liu2019poolnet}, have achieved very promising results. The SOD task defines saliency in an absolute way, where ground truth (GT) saliency maps are binary, \ie pixels belonging to salient objects are labeled as 1 while background pixels are labeled as 0. As a result, previous models learn to detect all salient objects in an image equally, without explicitly differentiating their different degrees of saliency.
	
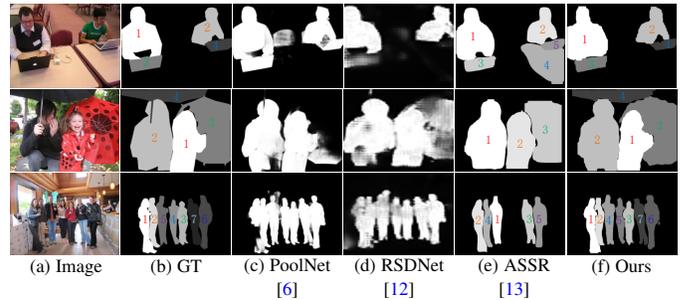
\begin{figure}[t]
		\graphicspath{{Figures/intro/}}
		\centering
		%\begin{overpic}[width=1\linewidth,grid,tics=2]{intro.pdf}
		\begin{overpic}[width=1\linewidth]{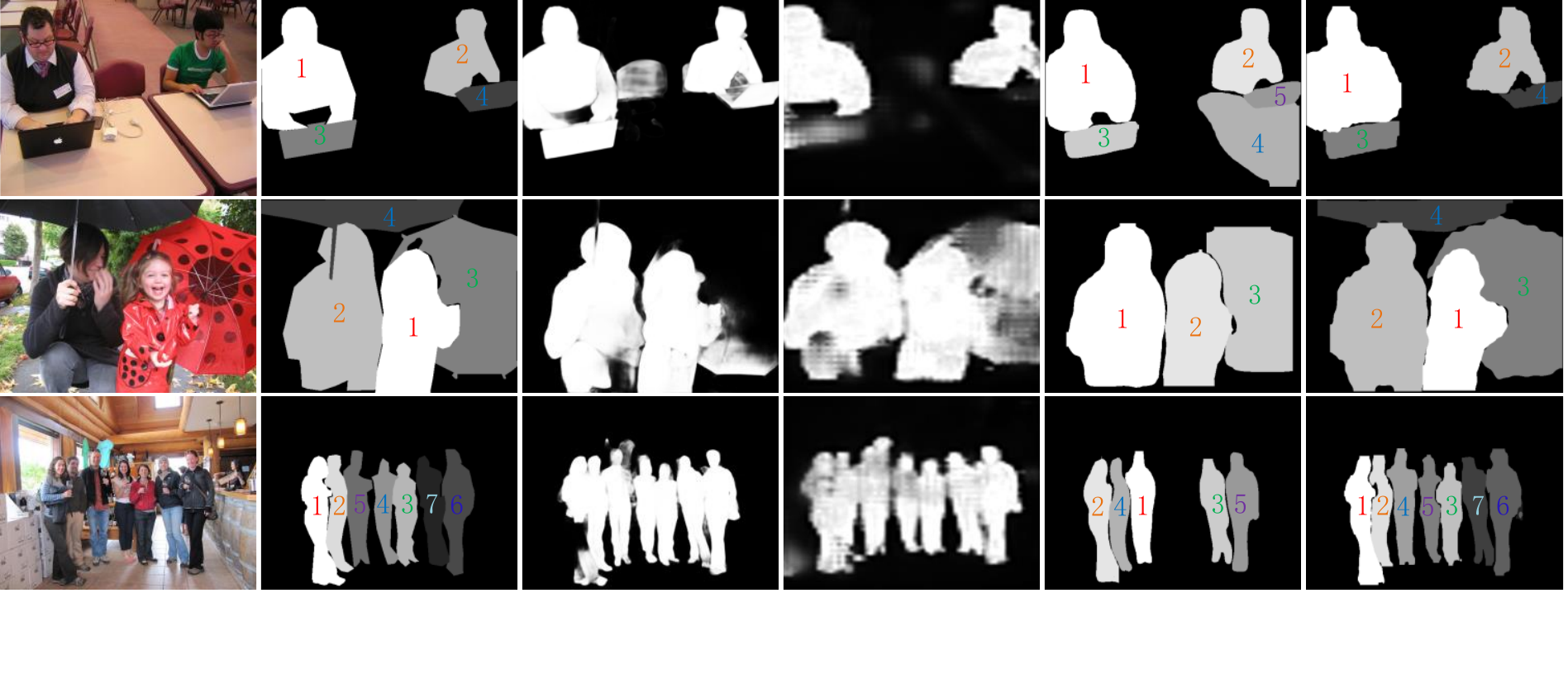}
			\put(3,3.5){\scriptsize (a) Image}
			\put(21,3.5){\scriptsize (b) GT}
			%		\put(33,3){\scriptsize (c) PoolNet \\ \cite{liu2019poolnet}}
			\put(35,0){\scriptsize \shortstack[c] {(c) PoolNet\\ \cite{liu2019poolnet}}}
			\put(51.5,0){\scriptsize \shortstack[c] {(d) RSDNet\\ \cite{amirul2018rsd}}}
			\put(70,0){\scriptsize \shortstack[c] {(e) ASSR\\ \cite{siris2020inferring}}}
			\put(87,3.5){\scriptsize (f) Ours}
		\end{overpic}
		\vspace{-7mm}
		\caption{\textbf{Comparison of different saliency models.} Salient object detection models highlight all salient objects equally in each image, as shown in column (c). In this paper, we aim at detecting instance-level relative saliency ranking, which assigns different saliency values to different salient objects to denote their degrees of saliency. (a) and (b): Three example images and their corresponding GT saliency maps. (d): In \cite{amirul2018rsd}, the authors predicted pixel-wise saliency ranking, without actually differentiating object instances. (e): In \cite{siris2020inferring}, saliency ranking is inferred based on attention shift, and only less than five objects are considered. (f): Our model shows more accurate and practical results for relative saliency ranking. To facilitate the discrimination of different rank orders, we also mark the rank orders on each salient instance.}
		\label{figure1}
		\vspace{-4mm}
	\end{figure}

	However, in a visual scene, different objects usually have noticeably different degrees of saliency. As human beings, it is easy for us to judge whether one object is more salient than another, which is known as relative saliency. Compared with the absolute saliency modeling in previous works, employing relative saliency is closer to the human visual attention mechanism, in which multiple stimuli will cause visual cortex neurons to compete with one another \cite{crick1990towards,motter1993focal,niebur1993oscillation,liu2018dsclrcn}. Some examples are shown in Figure~\ref{figure1}, where column (b) shows the relative saliency maps. Note that the lighter the mask, the more salient the object. Predicting relative saliency is also more practical since it enables us to clearly distinguish which object is the most salient and know the relative importance among objects, thus benefiting higher-level image understanding \cite{bylinskii2016should}. This can thus push the visual saliency research field forward. Although current absolute saliency models are somewhat able to predict relative saliency by generating different saliency values for different image regions, their lack of explicit learning makes their results inaccurate and incomplete, as shown in column (c) of Figure~\ref{figure1}.
	% However, the perception of saliency is a highly subjective task. Whether an object/region is salient or not shows noticeable heterogeneity for different individuals \cite{xu2018personalized}. This is also reflected in the annotation contradiction in salient object detection datasets, \eg, some images appear in both SOD \cite{movahedi2010sod} and ECSSD \cite{yan2013hs} datasets but show different annotations, and some datasets (\eg, DUT-O \cite{yang2013gbmr} and PASCAL-S \cite{li2014secrets}) are apt to annotate some objects as salient despite that they are usually considered as backgrounds in other datasets. Considering predicting absolute saliency is somewhat indeterminate, we argue that one can turn to predict relative saliency, \ie judging whether an object is more salient than the other one. Such a task is closer to the human visual attention mechanism in which multiple stimuli will cause visual cortex neurons to compete with each other \cite{crick1990towards,motter1993focal,niebur1993oscillation}.
	%based on the differential sensitivity of neurons in cortical visual areas with the presence of multiple competing stimuli \cite{motter1993focal}.
	
	As a pioneering work, Islam \etal \cite{amirul2018rsd} proposed to explicitly predict relative saliency maps that have different saliency values at different pixels, as shown in column (d) of Figure~\ref{figure1}. Then, they ranked the salient objects in each image according to the mean saliency value within each GT instance mask. They defined this task as relative saliency ranking. We argue that their method is a pixel-wise relative saliency solution instead of an object-level one since it predicts pixel-wise saliency and does not differentiate object instances. They have to use GT instance segmentation maps to obtain object-level relative saliency ranking results, which is infeasible for practical usage.
	%In addition, they regress pixel-wise relative saliency values using a convolution-deconvolution network, without considering high-level interactions among different salient objects and the context. As a result, their saliency maps often show ranking errors, as can be seen in column (d) of Figure~\ref{figure1}.
	In \cite{siris2020inferring}, Siris \etal proposed to predict saliency ranking by inferring human attention shift, which describes how humans sequentially select and shift attention from one object to another.
	%They first adopted Mask R-CNN \cite{he2017mask} to segment object instances in each image. Then, they built a deep model by combining scene-object relation attention and spatial attributes of objects to infer saliency rank. 
	However, the degree of saliency of an object mainly depends on the duration of gaze instead of the sequential order in which objects are focused. Thus, attention shift is largely different from relative saliency and instead very close to the scanpath prediction task 
	%\cite{xia2019predicting,xia2020evaluation}.
	\cite{xia2020evaluation}. 
	Hence, the saliency ranking of this method has a different mechanism and applications, as shown in Figure~\ref{figure1}, column (e).
	%Furthermore, their model only considers object-scene relation attention and spatial attributes of objects to infer saliency rank, preventing it from effectively performing complicated high-level reasoning for the SOD task. Figure~\ref{figure1}, column (e) shows the example results.
	
	In this paper, we follow \cite{amirul2018rsd} to infer relative salient object ranking. 
	%	Different from them, however, we directly perform saliency ranking at an object-level and present a unified model as the first end-to-end solution.
	\rev{The challenge of this problem mainly lies in two aspects. First, how to model the whole problem. \cite{amirul2018rsd} modeled it as a pixel-wise regression problem and trained a deep network using pixel-wise euclidean regression loss. Such a way leads to unsatisfactory results for both segmentation and ranking since pixel-wise training can not guarantee instance structures and using the regression scheme can not obtain accurate ranking order. In contrast, we follow \cite{siris2020inferring} and adopt a two-stage scheme to segment salient instances first and then rank them. Specifically, we improve the state-of-the-art instance segmentation model Mask R-CNN \cite{he2017mask} for salient instance segmentation. Subsequently, we add a saliency ranking branch to rank the relative saliency of each segmented instance. This scheme guarantees the segmentation result and focuses on the saliency ranking problem. \cite{siris2020inferring} used instance classification as the scapegoat of the saliency ranking problem to train its model, which is suboptimal since it can not directly optimize the original saliency ranking problem. Furthermore, dosing so requires fixed instance numbers, which is inflexible for practical usage. Hence, we propose to directly optimize the saliency ranking order using an improved ranking loss \cite{chen2016single}, which raises and suppresses the scores of highly-ranked and lowly-ranked instances, respectively, thus being more suitable and effective for this new task than the conventional losses used in \cite{amirul2018rsd,siris2020inferring}.}
	
	\rev{The second challenge is how to effectively infer relative saliency. \cite{amirul2018rsd} directly regressed relative saliency using a convolution-deconvolution network, which does not leverage any explicit saliency cues. \cite{siris2020inferring} considered the object-scene relation attention and spatial attributes of objects. However, more important saliency cues are ignored in their model. In this work, inspired by the biological evidence in \cite{crick1990towards,motter1993focal,niebur1993oscillation}, we consider the stimuli interaction and competition relation among different instances as the most important cue for inferring relative saliency. However, this can not be explicitly modeled in traditional CNNs. Hence, we build a graph reasoning model to achieve this goal using graph neural networks (GNNs) \cite{scarselli2008graph}, which are powerful tools to model relationships and propagate context information among nodes, thus being naturally suitable for our problem. Furthermore, considering that the saliency of a region has been found to be highly related to its local contrast, global contrast \cite{liu2016learning,li2015mdf}, and high-level semantic priors \cite{liu2016learning}, we propose to construct three more graphs to model these saliency factors. On the one hand, doing so leads to a unified GNN-based saliency ranking model. On the other hand, using GNNs for contrast inference is reasonable and straightforward. Different from previous works \cite{liu2016learning,li2015mdf} that combine the center feature and context features via concatenation and fully-connected (FC) layers, we treat the center instance and the context as graph nodes and use GNNs to mine contrast relationships among them, which is a both new and more powerful way.}
	
	%	Inspired by the biological evidence in \cite{crick1990towards,motter1993focal,niebur1993oscillation}, we first build a graph reasoning model for propagating stimuli interaction and competition relationships among different instances using graph neural networks (GNNs). Additionally, considering that the saliency of a region has been found to be highly related to its local contrast, global contrast \cite{liu2016learning,li2015mdf}, and high-level semantic priors \cite{liu2016learning}, we propose to construct three more graphs to model these saliency factors.
	%%	by incorporating local context, global context, and the person prior. 
	%	We also present an improved ranking loss \cite{chen2016single} to train our saliency ranking branch. It  in each sampled instance pair, thus being more suitable and effective for this new task than the conventional losses used in previous works.
	
	Moreover, existing datasets and evaluation metrics are also unsatisfactory for this new problem setting and practical usage. The PASCAL-S dataset \cite{li2014secrets} used in \cite{amirul2018rsd} has small-scale images. The COCO-SalRank dataset \cite{kalash2019rsd} uses sophisticated human-designed rules to determine salient instances, which cannot guarantee annotation accuracy. The dataset proposed in \cite{siris2020inferring} mainly focuses on attention shift and generates annotations for a fixed number of objects in each image instead of differentiating salient and non-salient objects, hence being unsuitable for both the relative saliency ranking task and practical usage. To this end, we build a new relative saliency ranking dataset with large-scale images and human-annotated salient objects. \rev{As for the evaluation metrics, the salient object ranking (SOR) metrics used in \cite{amirul2018rsd,kalash2019rsd,siris2020inferring} mainly focus on ranking performance, entirely or partially ignoring salient object detection and segmentation performance.} Therefore, we propose an improved SOR metric that comprehensively incorporates evaluations of all three types of performance. Our proposed dataset and evaluation metric are more suitable for this new task and can highly benefit future research.
	
	Finally, experimental results demonstrate that the saliency factors considered in this paper are more appropriate than those in previous works, and our proposed graph model and loss are much more effective for relative saliency ranking inference. As a result, our overall model performs favorably against previous methods when compared on both our dataset and existing ones. We also show a successful application of our proposed model on the adaptive image retargeting problem to demonstrate its practical usage.
	
	To sum up, the contributions of this work are as follows:
	\begin{compactitem}
		\item We propose the first end-to-end solution for the relative salient object ranking problem by adding a saliency ranking inference branch to a salient instance segmentation model.
		\item We present a novel graph neural reasoning module to simultaneously incorporate the stimuli interaction and competition relations, local contrast, global contrast, and high-level person prior for saliency ranking inference. A new ranking loss is also introduced to train this module.
		\item In contrast to previous saliency ranking datasets, we construct a new large-scale dataset with human-annotated salient objects, thus providing more accurate training and testing data.
		\item Considering the limitations of the previously used evaluation metric, we propose a new one for this task, which comprehensively considers salient instance detection, segmentation, and ranking performance.
		\item Experimental results verify the effectiveness of our saliency cues and the proposed model. We also show that our model outperforms previous methods on all available datasets and demonstrate its application to the adaptive image retargeting task.
	\end{compactitem}

	\section{Related Work}
	\subsection{Salient Object Detection}
	%Salient object detection (SOD) aims at localizing the most salient objects in each image and integrally segment them out. 
	Since the early efforts of \cite{liu2010learning}, SOD has become a more and more important research topic in the computer vision community. Early works \cite{liu2010learning,cheng2014gc} typically employed the contrast mechanism and some other prior knowledge
	%, such as background prior \cite{wei2012gs,yang2013gbmr}, compactness prior \cite{perazzi2012sf}, and objectness prior \cite{chang2011fusing},
	to model visual saliency.
	
	In recent years, many works have introduced deep neural networks (DNNs) to the SOD task and have achieved significant performance improvements. Some models \cite{li2015mdf,zhao2015mcdl,wang2015legs} treat salient object detection as a pixel/superpixel-wise classification problem and run deep classification models on multi-scale regions. %However, such models are very computationally expensive.
	Later works, such as \cite{kuen2016recurrent,wang2016rfcn,Wang2017srm}, usually adopt fully convolutional networks (FCNs) \cite{long2015fcn} to improve the computational efficiency and preserve spatial structures. More recent methods also consider fusing CNN features of different layers to better refine local structures and obtain finer saliency maps. Two network styles are often used. One is the HED \cite{xie2015hed} style which interpolates multilayer feature maps to the same size and fuses them once, such as \cite{hou2017dss}. The other is the U-Net \cite{ronneberger2015unet} style which progressively fuses feature maps from deep layers to shallow layers in a top-down manner, such as \cite{liu2016dhsnet,luo2017nldf}. Furthermore, other models combine various complementary information to enhance the model performance, such as the attention mechanism \cite{liu2018picanet,chen2018ra,wang2019salient}, contour detection \cite{liu2019poolnet,wang2019salient,zhou2020interactive}, and gaze prediction \cite{wang2019inferring}.
	For a more comprehensive survey please refer to \cite{wang2019sodsurvey}. Although some SOD models can also predict different saliency values for different pixels, they are originally designed for binary SOD and focus more on accurately segmenting out salient objects, hence performing unsatisfactorily for saliency ranking detection.
	%Furthermore, none of the previous deep SOD models consider explicitly model contrast and top-down semantic priors. In contrast, we use graph networks to explicitly model instance interaction relations, local and global contrasts, and semantic priors to infer saliency ranking.
	
	\subsection{Eye Fixation Prediction}
	Eye fixation prediction (EFP) was the first and remains a very important school of saliency detection. Different from SOD, it only aims to localize the most salient regions in each image. Traditional methods typically incorporate local contrast \cite{itti1998model}, global contrast \cite{borji2012exploiting}, and high-level semantics \cite{cerf2007predicting,judd2009learning,borji2012boosting} to infer saliency. In recent years, CNNs have been widely adopted for this task and have largely improved EFP performance. Most models directly treat EFP as a per-pixel classification task and adopt FCN-style models for each image \cite{huang2015salicon,kruthiventi2017deepfix,wang2017dva,cornia2018predicting,kroner2020contextual}. Besides, long-short term memory (LSTM) \cite{hochreiter1997long} and generative adversarial networks (GANs) \cite{goodfellow2014gan} have also been introduced into the EFP task to iteratively refine saliency maps \cite{cornia2018predicting} and incorporate data-driven saliency losses \cite{pan2017salgan,che2019gaze}. Other models explicitly model saliency cues in deep networks, such as the local and global contrast in \cite{liu2016learning}, global stimuli competition and scene modulation in \cite{liu2018dsclrcn}, and emotional properties in \cite{fan2018emotional}. In this work, we share similar ideas with several EFP models to infer saliency from local and global contrasts, stimuli competition, and high-level semantic priors. The differences are i) We focus on inferring saliency ranking for salient instances using the proposed ranking loss instead of directly conducting saliency classification as done in most EFP models; ii) We construct a graph reasoning module to effectively model and combine different saliency cues.
	
	\subsection{Salient Instance Segmentation}
	Since, in this paper, we aim at simultaneously segmenting salient instances and inferring their saliency ranking, a highly related topic is salient instance segmentation (SIS), which is also a relatively new topic. Li \etal \cite{li2017ilso} were the first to introduce the instance-level salient object segmentation task and constructed the first dataset with pixel-wise SIS annotations. They also proposed an SIS model by combining saliency and contour maps predicted by a CNN.
	%They also propose an SIS model name MSRNet. Specifically, they use a multiscale CNN to simultaneously predict a saliency map and a contour map for each image. Then object proposals can be obtained by applying the MCG \cite{arbelaez2014mcg} algorithm and a subset optimization method on the contour map. Finally, a CRF model is used to generate the SIS results from the predicted saliency map and the proposals.
	
	Since then, other researchers have mainly adapted and improved state-of-the-art instance segmentation models for the SIS task. Kampffmeyer \etal \cite{kampffmeyer2018connnet} incorporated their proposed pixel-pair-based connectivity formulation into Mask R-CNN \cite{he2017mask} and achieved improved performance. In \cite{fan2019s4net}, Fan \etal also built a new SIS model based on Mask R-CNN and presented a novel RoIMasking layer to additionally consider surrounding background context for training the mask head. Tian \etal \cite{tian2020weakly} proposed to combine saliency detection, boundary detection, and centroid detection to perform weakly-supervised SIS from class and subitizing labels.
	
	In this paper, we improve Mask R-CNN using techniques from \cite{liu2018panet} to segment salient instances and find that our model can achieve better SIS performance. Then, we add a saliency ranking branch to it to further infer relative saliency ranking for segmented instances.
	
	\subsection{Graph Neural Network}
	Graphs use nodes and edges to represent targets and their relationships. In recent years, researchers have proposed GNNs \cite{scarselli2008graph} to extend neural networks for modeling data structures in graph domains, such as social networks, molecular graphs, knowledge graphs, texts, etc. Very recently, GNNs have also been explored for vision tasks. Wang \etal \cite{wang2018videos} represented videos as space-time region graphs and used graph convolutional networks (GCNs) to capture similarity and spatial-temporal relations for action recognition. In \cite{huang2019dynamic}, two dynamic graphs were constructed to capture appearance/motion changes and relative position changes among objects for activity recognition. Chen \etal \cite{chen2018iterative} built three global graphs for iterative visual reasoning in object detection, where a knowledge graph, a region graph, and an assignment graph were used for semantic relation propagation, spatial relation propagation, and region classification, respectively. Similarly, GNNs are widely used for image captioning \cite{yao2018exploring}, visual question answering \cite{li2019relation}, action forecasting \cite{sun2019relational}, video object segmentation \cite{wang2019zero,lu2020video}, RGB-D saliency detection \cite{luo2020cascade}, and so on, by modeling various relations.
	%, such as semantic relationships and spatial ones. 
	In this work, we are the first to investigate GNN models for saliency ranking inference by using four graphs to simultaneously model four related relationships.

	\section{Proposed Dataset}
	
\begin{figure*}[!t]
		\graphicspath{{Figures/dataset/}}
		\centering
		\includegraphics[width=1\linewidth]{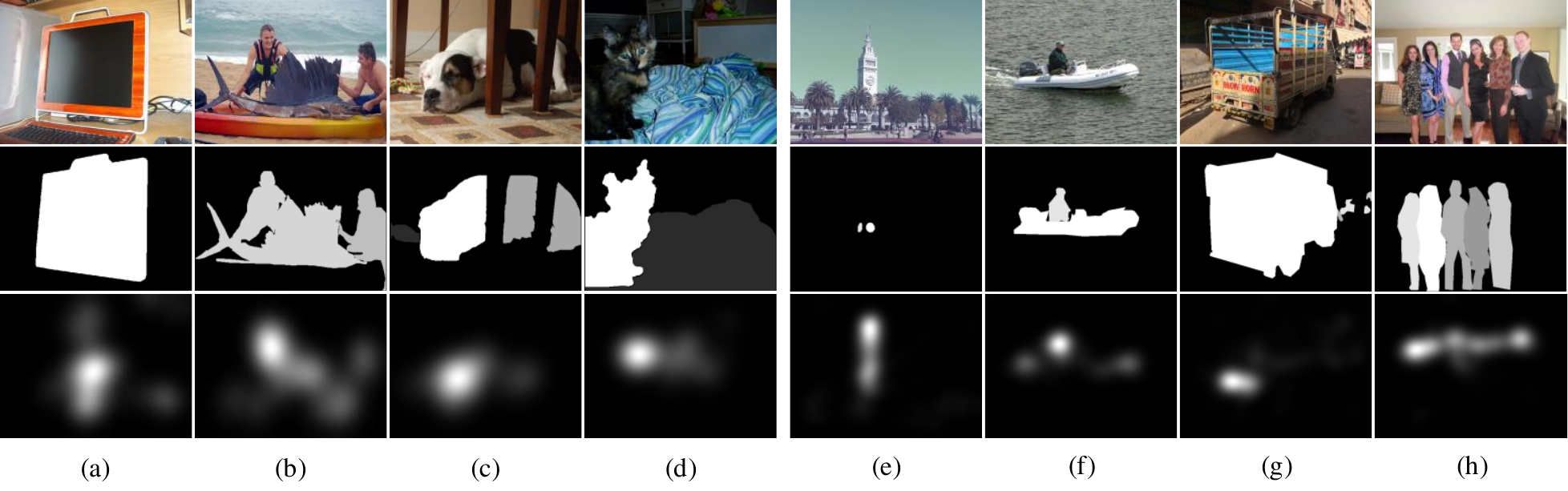}
		\caption{\textbf{Limitations of previous saliency ranking datasets}. We show the limitations of the PASCAL-S dataset \cite{li2014secrets} in the first four columns and those of the Siris' dataset \cite{siris2020inferring} in the last four columns. The second and the third row indicate their saliency ranking annotations and the eye gaze saliency maps, respectively.}
		\vspace{-0.2cm}
		\label{fig:prev_dataset_limit}
		\vspace{-0.3cm}
	\end{figure*}

	In \cite{amirul2018rsd}, Islam \etal conducted model training and evaluation on the PASCAL-S \cite{li2014secrets} dataset for relative saliency ranking detection. However, this dataset is not the ideal choice for this task, for several reasons. First, it only contains a total of 850 images, 40.4\% of which have only one saliency rank in the image, \rev{as shown in Figure~\ref{fig:prev_dataset_limit}(a)}. These images cannot be used for training and evaluating saliency ranking models, \rev{while} the remaining images are too few for training deep models. Second, there are two problems occur in many images of this dataset. The first is that multiple instances in the same image are annotated with the same rank, \rev{as shown in Figure~\ref{fig:prev_dataset_limit}(b)}. The other is that some instances are over-segmented into multiple regions with different rank values, \rev{such as the dog in Figure~\ref{fig:prev_dataset_limit}(c)}. Both cases are inappropriate for the instance-level saliency ranking detection task. Third, some of the annotated salient objects in this dataset actually belong to ``stuff'' classes, which are amorphous regions of similar texture or material, such as trees, mountains, \rev{and the quilt in Figure~\ref{fig:prev_dataset_limit}(d)}. These classes are usually uncountable and it is difficult to define ``instances'' for them.
	
	To this end, in both \cite{kalash2019rsd} and \cite{siris2020inferring}, the authors constructed new saliency ranking benchmark datasets by leveraging the existing MS-COCO \cite{lin2014coco} dataset and the SALICON \cite{jiang2015salicon} dataset, respectively. The former has more than 100k images with 80-classes of pixel-wise instance mask annotations. The latter contains 10,000 training and 5,000 validation images with gaze data obtained by mouse tracking, with all images sampled from MS-COCO. Thus, intuitively, it should be easy to use the gaze data from SALICON to rank the object instances from MS-COCO. However, this is non-trivial since there are several challenges in combining the two datasets. The first is that, since MS-COCO only has 80 annotated classes of object instances, several images have object instances with adequate fixations but no mask annotations. The second challenge is that many COCO images have dozens of annotated instances, while we cannot use all of them for saliency ranking. This is simply because saliency perception is very subjective and it is difficult even for humans to undoubtedly rank the saliency of many objects, especially, \rev{objects with less degree of saliency}. Hence, certain salient instances should be selected from the complex background, while non-salient ones must be ignored. The third challenge is that there are several annotation errors in MS-COCO, such as over or under-segmented images. All of these challenges hinder the direct usage of these two datasets.
	
	%加一个不同rank的rank 误差的分析，计算8个GT rank对应的预测rank与GT的绝对值差。应该是GT rank越高的，误差越小，GT rank越低的误差越大
	\begin{table*} [!t]
		\begin{center}
			\caption{\textbf{Statistical comparison of three saliency ranking datasets.} We provide a comparison in terms of the total image numbers and the image distributions over different salient instance numbers.}
			%\vspace{-0.2cm}
			\label{soc-r_stat}
			\footnotesize
			\begin{tabular}{@{}L{2cm}|C{1cm}|R{0.6cm}R{0.6cm}R{0.6cm}R{0.6cm}R{0.6cm}R{0.6cm}R{0.6cm}R{0.6cm}R{0.6cm}}
				\toprule
				\multirow{2}{*}{Datasets} &\multirow{2}{*}{\makecell[c]{Number\\ of Images}} & \multicolumn{9}{c}{Percentages of Images with Different Salient Instance Numbers (\%)}   \\ \cmidrule{3-11}
				&                                              &1  &2   &3     &4     &5     &6     &7     &8   &9+
				\\  \midrule
				PASCAL-S \cite{li2014secrets}     & 850   & 35.3  & 26.7  & 15.9  & 8.6   & 5.1   & 3.3   & 2.1   & 1.3   & 1.8
				\\
				Siris' \cite{siris2020inferring} & 11,500 &  -    & 9.8   & 14.0  & 12.5  & 63.6  &   -   &   -   &   -   &-
				\\  \midrule
				Ours                              & 8,988  &  -    & 34.1  & 29.9  & 17.5  & 9.4   & 5.0   & 2.5   & 1.7   &-
				\\
				\bottomrule
			\end{tabular}
			\vspace{-0.3cm}
		\end{center}{}
	\end{table*}
	
	Considering the above problems, in \cite{kalash2019rsd}, Kalash \etal used complicated hand-designed rules with adjustable parameters to filter out inappropriate images and select salient object instances when building their datasets. By using loose and strict parameters, they constructed a noisy and clean dataset, respectively. Although they tuned the parameters carefully and verified them on smaller-scale data, their hand-designed rules cannot fit all scenarios, especially for complex visual scenes. In \cite{siris2020inferring}, Siris \etal simply filtered out images without object annotations, as well as those containing smaller objects completely enclosed by larger ones. This simple pre-processing introduced significant noise to the selected images, however, especially since they could not filter out images with salient objects outside the 80 MS-COCO classes, \rev{such as the building in Figure~\ref{fig:prev_dataset_limit}(e)}. Besides, they generated the saliency ranking annotations according to the attention shift, making their dataset \rev{very close to the scanpath prediction task rather than the relative saliency ranking task}. \rev{For example, they ranked the boat in Figure~\ref{fig:prev_dataset_limit}(f) as the most salient because it's in the center of the image, at which people usually look first, although the person is much more salient than the boat according to the eye gaze study shown in the third row}. Last but not least, unlike \cite{kalash2019rsd} and our dataset, they did not define salient and non-salient objects in each image. \rev{Their dataset simply ranks the five most salient instances, thus including some non-salient objects as shown in Figure~\ref{fig:prev_dataset_limit}(g) or ignoring some salient objects as shown in Figure~\ref{fig:prev_dataset_limit}(h). Hence, it is not only less reasonable in defining salient objects, but also inflexible in practical usage.}
	
	In this paper, we construct a more accurately annotated dataset for the relative salient instance ranking task. We first find the 15,000 SALICON images from the MS-COCO dataset and extract their instance segmentation masks. Then, we show each image with their instance annotations to different subjects and ask them to pick out appropriate images and select salient objects. The subjects include five postgraduate students aging from 20 to 30, with four males and one female. The image and salient object selection rules are as follows: \textbf{i)} For each image shown, each subject selects the objects he/she thinks are salient. \textbf{ii)} If an image has a salient object not shown in the MS-COCO instance annotations, it should be marked as inappropriate. \textbf{iii)} If an image has obvious segmentation errors, \eg having over or under-segmented instances, it should be marked as inappropriate. \textbf{iv)} If an image has more than eight or less than two salient instances, or does not have obvious salient objects, it should be marked as inappropriate. 
	%we limit each image should have more than 2 salient instances since we need to conduct instance ranking and 
	We limit the maximum number of salient instances to eight following the PASCAL-S dataset, which has at most seven saliency ranks in each image, and considering the relatively large number of objects in MS-COCO images. After manual selection and annotation, we filter out the images marked as inappropriate by more than three subjects. Similarly, for the remaining images, we label the objects annotated as salient by more than three subjects as salient objects. Then, for saliency ranking within each image, we follow \cite{kalash2019rsd} and leverage the saliency maps provided by the SALICON dataset instead of the fixation points. However, rather than using the average saliency value within each instance,
	%normalized by its size
	we rank the labeled salient objects according to the maximum saliency value within each instance mask, since the degree of saliency of an object is mainly determined by its distinctive parts.
	%  \cite{yarbus2013eye} We also remove images having multiple instances with the same rank.
	
\begin{figure}[!t]
		\graphicspath{{Figures/dataset/}}
		\centering
		\includegraphics[width=1\linewidth]{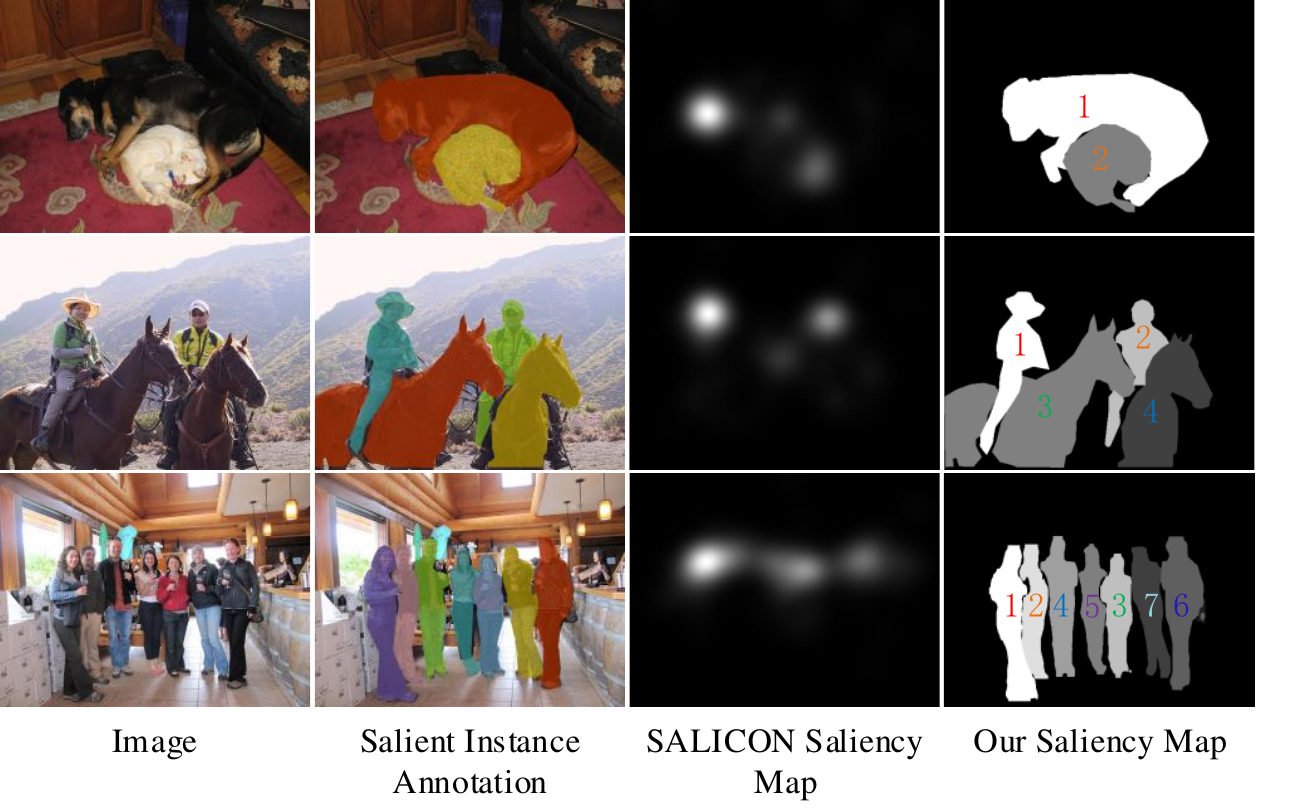}
		%\begin{overpic}[width=1\linewidth,grid,tics=1]{PiCANet.pdf}
		%\begin{overpic}[width=1\linewidth]{S2MA.pdf}
		%\end{overpic}
		\caption{\textbf{Examples of our proposed dataset.} For each image, we show our annotated salient instances and the saliency maps from the SALICON \cite{jiang2015salicon} dataset. Based on these two kinds of annotations, we generate relative saliency ranking maps, as shown in the last column.}
		\label{dataset_fig}
		\vspace{-0.3cm}
	\end{figure}

	Finally, we obtain a total of 8,988 images, which are divided into 6,059 training images and 2,929 test images based on the training and validation split of SALICON. Similar to the PASCAL-S dataset, we visualize both instance segmentation and relative saliency ranks as saliency maps. However, different from PASCAL-S, for each image we assign the saliency values in different salient instance masks by uniformly dividing [0,255] by their saliency rank orders. Figure~\ref{dataset_fig} shows three example images from our proposed dataset with annotated salient instances, SALICON saliency maps, and the generated saliency ranking maps.
	
	\textbf{Statistical analysis.} Table~\ref{soc-r_stat} shows a statistical comparison between our proposed dataset, PASCAL-S \cite{li2014secrets}, and Siris' dataset \cite{siris2020inferring}.\footnote[1]{The authors of \cite{kalash2019rsd} have not released their proposed COCO-SalRank dataset.} As can be seen, our dataset has large-scale image data, similar to Siris' dataset. PASCAL-S has many images with only one salient instance, which cannot be used for saliency ranking. Siris' dataset has at most five salient instances and the images with exactly five instances have a large percentage of 63.6\%, hence deviating from real application scenarios. In contrast, our dataset is more balanced. Many images have two or three salient instances and more than 30\% of images have four or more. It is worth noting we find that about 9\% of images have more than five salient instances according to human annotations. Hence, compared with the existing two datasets \cite{kalash2019rsd,siris2020inferring}, it is necessary for our dataset to consider a larger salient instance number limitation, and the findings also demonstrate the difficulty of our dataset.
	
	\begin{figure*}[!t]
		\graphicspath{{Figures/category_stat/}}
		\centering
		\includegraphics[width=1\linewidth]{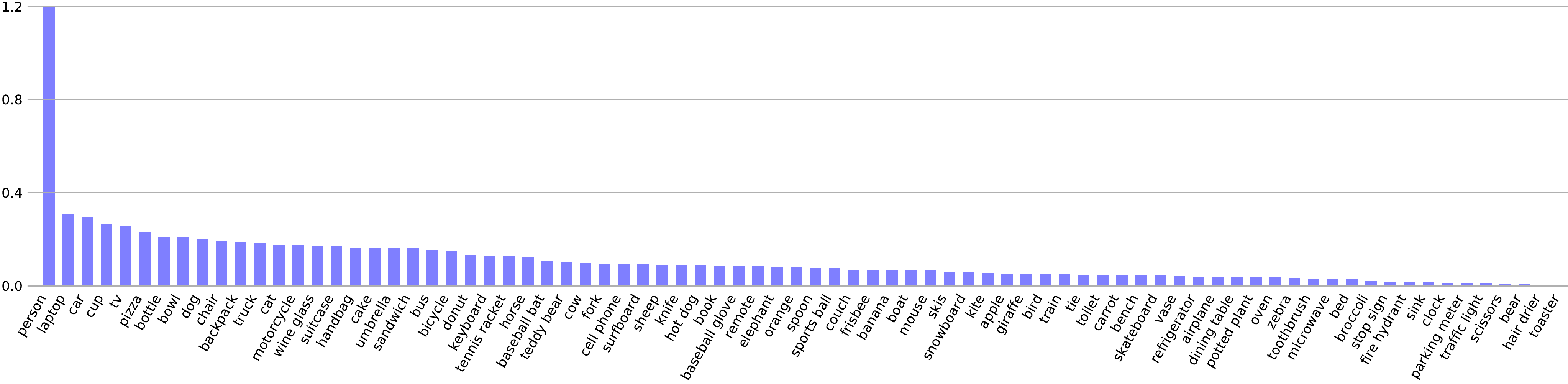}
		%\begin{overpic}[width=1\linewidth,grid,tics=1]{PiCANet.pdf}
		%\begin{overpic}[width=1\linewidth]{S2MA.pdf}
		%\end{overpic}
		\caption{\textbf{Comparison of the average saliency scores among 80 COCO categories on our proposed dataset.} For better visualization, we plot the histograms of $log(1+S_j)$ for each category $j$.}
		\label{fig:category_score}
		\vspace{-0.1cm}
	\end{figure*}
	
	\begin{figure*}[!t]
		\graphicspath{{Figures/network/}}
		\centering
		\includegraphics[width=1\linewidth]{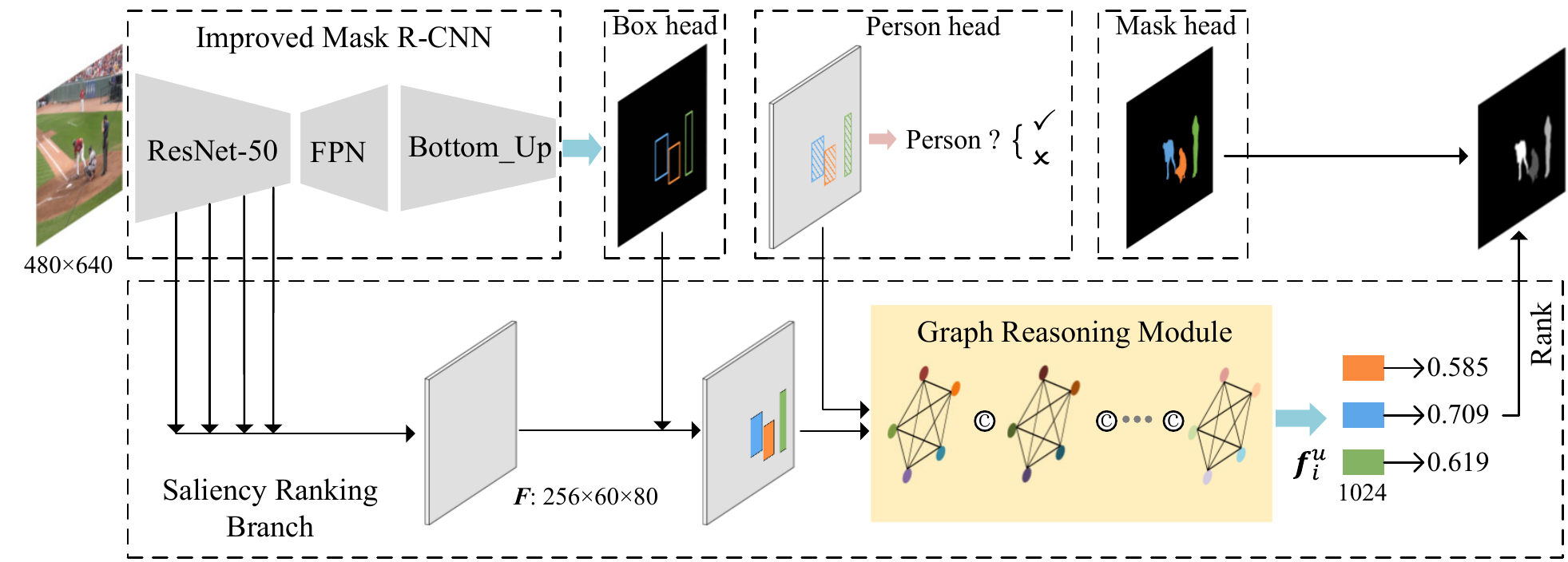}
		\caption{\textbf{Network architecture of our proposed model.} We first use our improved Mask R-CNN to obtain salient instance segmentation results. A person head is also added in parallel with the box and mask heads to estimate each instance as being a person or not. Then, we fuse ResNet features to get a fused saliency feature map $\bm{F}$, from which we can extract instance features, as well as local and global context features for saliency ranking inference. We build a graph reasoning model to incorporate instance interaction relations, local and global contrast inference, and the person prior knowledge, hence obtaining the updated instance features $\bm{f}_i^u$. Finally, we predict the saliency scores for the salient instances and rank them to obtain the final saliency ranking map.}
		\label{network_fig}
		\vspace{-0.3cm}
	\end{figure*}
	
	Considering that the degree of saliency is also largely influenced by the object category information, \eg faces, texts, cars, and persons have been found to be more salient than other objects in \cite{judd2009learning,borji2012boosting,liu2016learning}, we also investigate the average degree of saliency of different object categories on our proposed dataset. Specifically, for each rank $i$, where $i\in\{1,2,\cdots,8\}$ denotes the descending rank order, we count the proportion of each COCO category $j$ as $p_{i,j}$. Then, we assign an opposite score as $9-i$ for rank order $i$, hence assigning \rev{larger scores} for higher orders. Finally, the average saliency score of category $j$ can be calculated as:
	\begin{equation} \label{avg_score}
	S_j=\sum_{i=1}^{8}(9-i)p_{i,j}.
	\end{equation}
	We provide a visual comparison of the average saliency scores among the 80 COCO categories in Figure~\ref{fig:category_score}. As can be seen, the ``person" category is the most salient one and achieves much higher scores than other categories in our proposed dataset. This observation also matches the findings in \cite{judd2009learning,borji2012boosting,liu2016learning}.
	
	\section{Proposed Model}
	As shown in Figure~\ref{network_fig}, we first use an improved Mask R-CNN to segment salient instances for the input image and then add a branch to infer the saliency ranking for each instance. In this branch, we first construct a feature map from the network backbone. Then, we extract a saliency feature and a local context feature for each instance, obtaining appearance information, and a local contrast cue. At the same time, we also extract a set of global context features and a person-sensitive feature for incorporating global contrast and a person prior. Next, we adopt a graph reasoning module to propagate competition relationships among different instances and obtain updated features. Finally, we predict a saliency score for each instance and obtain the final saliency ranking map. The whole network works end-to-end.
	
	\subsection{Mask R-CNN for Salient Instance Segmentation}
	We improve Mask R-CNN with a ResNet-FPN backbone as our base model to segment salient instances, as shown in Figure~\ref{network_fig}. Specifically, the pretrained ResNet-50 \cite{he2016resnet} network is first used to extract convolutional (Conv) feature maps at four levels. The multi-level feature maps correspond to four residual blocks named ``Res2" to ``Res5",
	whose scales range from 1/4 to 1/32 of the size of the input image. Next, the FPN architecture \cite{lin2017fpn} is adopted to progressively fuse multi-level features in a top-down manner. In each FPN module, the previous FPN feature is upsampled and fused with a corresponding ResNet feature to obtain a new 256-D FPN feature. Another max-pooling layer is further deployed on the coarsest FPN feature to obtain a feature map with a scale of 1/64. As such, five-level FPN features are obtained with varying scales and strides. Then, following \cite{liu2018panet}, we add an extra bottom-up path to further propagate low-level information to high-level feature maps and also shorten the information path between lower layers and the topmost feature. As a result, each of the multi-level features can incorporate both lower-level and higher-level information.
	
	Next, region proposal networks (RPNs) \cite{ren2015faster} are used to generate salient object proposals. Each of the five-level features has a specific RPN to generate proposals with an adaptive scale.
	%where large proposals are generated at FPN levels with large strides and small proposals are generated at FPN levels with small strides.
	Subsequently, an RoI-specific feature can be extracted for each RoI by RoIAlign \cite{he2017mask} and is further inputted into a box head and a mask head. The former uses two FC layers to conduct saliency classification and box regression for each proposal. For the latter, we follow \cite{liu2018panet} and fuse a convolution-deconvolution branch and a parallel FC branch to obtain instance segmentation masks, which we find very useful for salient instance segmentation. Following the interleaved execution strategy in \cite{chen2019hybrid}, here we use the box coordinates refined by the box head to train the mask head, thus supplying it with more accurate bounding-boxes to promote its performance. Finally, we obtain segmented salient instances with their bounding-boxes and masks for each image.
	
	\subsection{Saliency Ranking Branch}
	We add a saliency ranking branch to the improved Mask R-CNN to infer the relative saliency ranking of the segmented salient instances. First, the feature maps of the four residual blocks of the ResNet-50 backbone are fused to obtain a saliency ranking feature map. To be specific, we transform each feature map to 256 channels and then interpolate it to the 1/8 scale, which is found to be appropriate for most instances. Then, we fuse the four feature maps by element-wise sum as the saliency ranking feature map $\bm{F}$. For each instance $i$, we use RoIAlign \cite{he2017mask} to extract a RoI feature with a spatial size of $7\times 7$ from $\bm{F}$ using the detected bounding-box. Then, we use two FC layers with 1024 channels and the ReLU activation function to obtain the instance feature $\bm{f}_i$. Finally, we construct a novel graph reasoning module to infer the relative saliency of each salient instance.
	
	%\subsubsection{Saliency Feature Extraction}
	%Next, we extract the saliency feature for each instance from $\bm{F}$, as shown in Figure~\ref{feature_fig}. First, for each instance $i$, we use RoIAlign \cite{he2017mask} to extract its RoI feature with the spatial size $7\times 7$ from its bounding-box. Then we use two FC layers with 1024 nodes and the ReLU activation function to obtain the appearance feature $\bm{f}_i^a$. Second, we consider incorporating the local context for local contrast learning. We enlarge the bounding-box of instance $i$ twice and then use RoIAlign to extract the local context RoI feature. Similarly, we use two FC layers to obtain the local context feature $\bm{f}_i^l$. Third, we also consider integrating the global context for learning global contrast. We first use max-pooling to pool $\bm{F}$ to the spatial size of $7\times 7$ and then obtain the global context feature $\bm{f}^g$ by FC layers. Last, considering that the spatial location and the object size of each instance are also very important for its saliency perception, we take the bounding-box coordinate information into account. Specifically, we use the positional encoding method in \cite{vaswani2017attention} to encode the box coordinates $[x_i,y_i,w_i,h_i]$ of instance $i$ into a 1024-d position feature $\bm{f}_i^p$ via sine and cosine functions. Finally, we concatenate the four features and then use an FC layer to generate the final saliency feature $\bm{f}_i$, which contains the appearance information, local, global contrasts, and the position information.
	
	\subsubsection{Interaction Relation Graph}
	
	\begin{figure}[h]
		\graphicspath{{Figures/graph_module/}}
		\centering
		\includegraphics[width=1\linewidth]{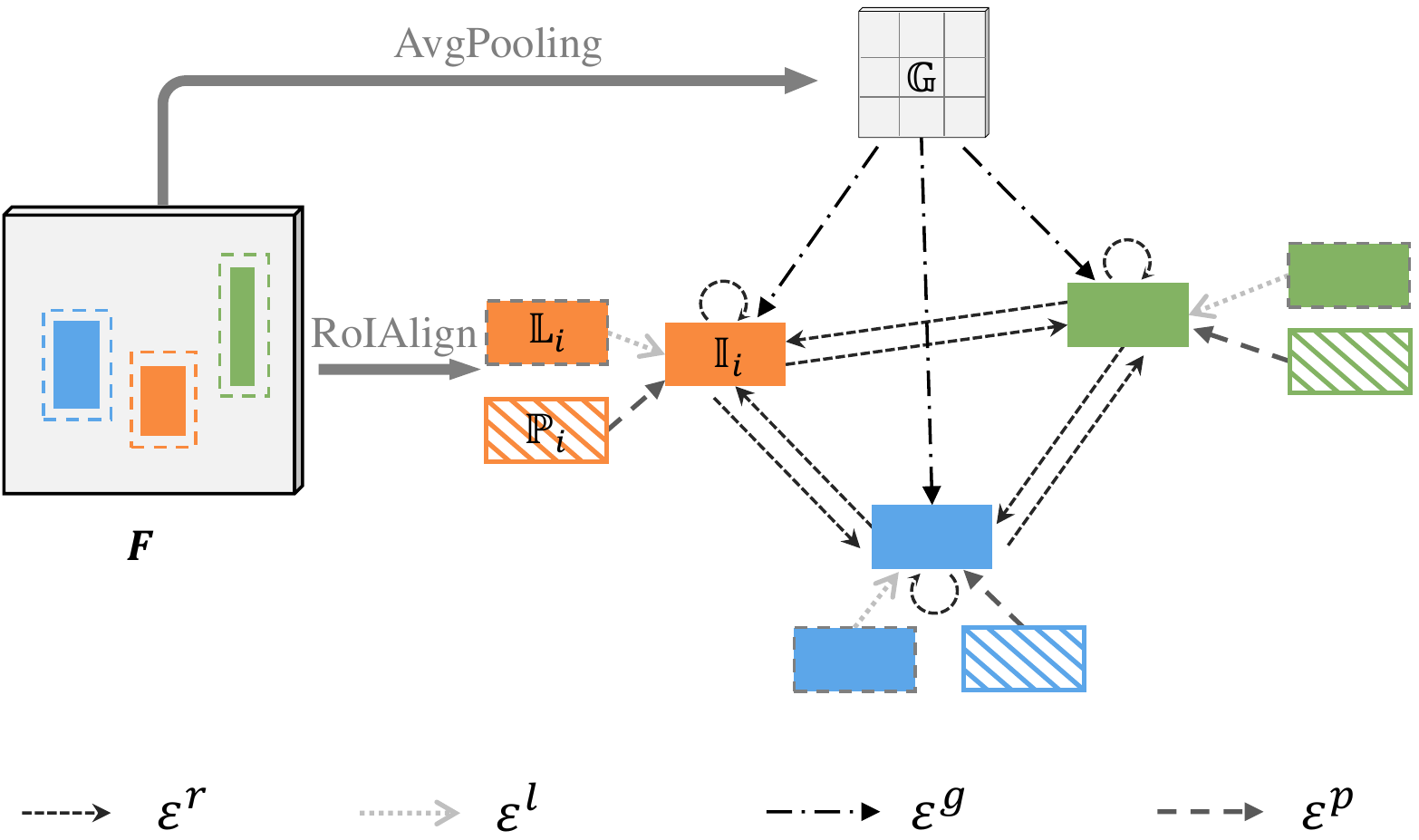}
		\caption{\textbf{Illustration of the our proposed graph reasoning module.} From the segmented salient instances and the saliency ranking feature map $\bm{F}$, we construct instance nodes $\{\mathbb{I}_i\}_{i=1}^{N}$, local context nodes $\{\mathbb{L}_i\}_{i=1}^{N}$, person prior nodes $\{\mathbb{P}_i\}_{i=1}^{N}$, and $M\times M$ global context nodes $\{\mathbb{G}_i\}_{i=1}^{M^2}$. Then, we build an interaction relation graph $\mathcal{G}^r$, a local contrast graph $\mathcal{G}^l$, a global contrast graph $\mathcal{G}^g$, and a person prior graph $\mathcal{G}^p$. In $\mathcal{G}^r$, each $\mathbb{I}_i$ is connected to each other and themselves. In $\mathcal{G}^l$ and $\mathcal{G}^p$, only $\mathbb{L}_i$ or $\mathbb{P}_i$ is connected to $\mathbb{I}_i$, respectively. In $\mathcal{G}^g$, all $\{\mathbb{G}_j\}$ are connected to each $\mathbb{I}_i$.}
		\label{fig:graph}
		\vspace{-0.3cm}
	\end{figure}
	
	A straightforward way to infer the saliency ranking of each instance is to directly regress a saliency value from its instance feature $\bm{f}_i$. However, inspired by the biological discovery that multiple stimuli can cause competition in the human visual attention system, we propose to capture the interaction relationships among different instances to promote saliency ranking inference performance. Specifically, for an image with $N$ instances, we denote each instance as $\mathbb{I}_i$, where $i\in\{1,2,\cdots,N\}$. We build an interaction relation graph $\mathcal{G}^r=(\mathcal{V}^r,\mathcal{E}^r)$, where $\mathcal{V}^r=\{\mathbb{I}_i\}_{i=1}^{N}$ is a set of nodes corresponding to the $N$ instances, while $\mathcal{E}^r$ is a set of interaction relation edges. We design $\mathcal{G}^r$ as a fully-connected graph, \ie each node is connected to all $N$ nodes, including itself with a self-connection edge, as shown in Figure~\ref{fig:graph}. As a result, $\mathcal{E}^r$ has $N\times N$ edges, where each edge $e_{i,j}^r$ denotes the connection from $\mathbb{I}_i$ to $\mathbb{I}_j$. By each self-connection edge $e_{i,i}^r$, we model how much information from $\mathbb{I}_i$ needs to be preserved in the graph updating process. Other edges are used to model the stimuli competition relationships in each instance pair.
	
	A GNN first uses an aggregator function to aggregate messages for each node from their neighboring nodes, and then updates their node representation using the collected message. Here we introduce an attention mechanism to dynamically modulate the message propagation at each edge. For instance $\mathbb{I}_i$ with feature $\bm{f}_i$, its relation neighbor $\mathcal{N}_i^r$ is exactly equal to the node set $\mathcal{V}^r$ in our fully-connected graph. Then we have the aggregator as:
	\begin{equation} \label{agg_r}
	\bm{h}_{\mathcal{N}_i^r}=\sum_{j=1}^N\alpha_{ij}^r\bm{W}_a^r\bm{f}_j,
	\end{equation}
	where $\bm{W}_a^r\in{\mathbb{R}^{C^r\times 1024}}$ projects the neighboring node features into the relation aggregating feature space with $C^r$ dimensions. The attention weight $\alpha_{ij}^r$ modulates the edge $e_{i,j}^r$ and it can be computed by:
	\begin{equation} \label{att_r}
	\alpha_{ij}^r=\frac{1}{N}\text{ReLU}((\bm{W}_{\alpha}^r)^{\top}(\bm{U}^r\bm{f}_i\lVert\bm{V}^r\bm{f}_j)),
	\end{equation}
	where $\bm{U}^r,\bm{V}^r\in{\mathbb{R}^{C^r\times 1024}}$ embed instance features into two attention subspaces with $C^r$ dimensions, $\lVert$ represents the concatenation operation, and $\bm{W}_{\alpha}^r\in{\mathbb{R}^{2C^r}}$ projects the concatenated feature into a scalar for attention computation.
	
	\subsubsection{Local Contrast Graph}
	Considering the effectiveness and importance of local contrast inference for saliency prediction, we also propose another local contrast graph $\mathcal{G}^l=(\mathcal{V}^l,\mathcal{E}^l)$ by incorporating local contexts. For each instance $\mathbb{I}_i$, we first extract its local context $\mathbb{L}_i$ by enlarging the bounding-box of $\mathbb{I}_i$ twice and then use RoIAlign to extract the RoI feature. Similarly, we use two FC layers to obtain the local context feature $\bm{f}_i^l$. Then, we include both instances and local contexts in $\mathcal{V}^l$, \ie $\mathcal{V}^l=\{\mathbb{I}_i, \mathbb{L}_i\}_{i=1}^{N}$. In $\mathcal{G}^l$, we only connect each local context to the corresponding instance to infer the local contrast of the latter, \ie $\mathcal{E}^l$ has $N$ edges and each edge $e_{i}^l$ corresponds the connection from $\mathbb{L}_i$ to $\mathbb{I}_i$. Hence, $\mathcal{G}^l$ forms a sparsely connected and directed graph. For $\mathbb{I}_i$, its local contrast neighbor $\mathcal{N}_i^l$ only includes $\mathbb{L}_i$. Hence, its aggregator function can be written as:
	\begin{equation} \label{agg_l}
	\bm{h}_{\mathcal{N}_i^l}=\alpha_{i}^l\bm{W}_a^l\bm{f}_i^l.
	\end{equation}
	Similarly, $\bm{W}_a^l\in{\mathbb{R}^{C^l\times 1024}}$ projects the local context feature into the local contrast feature space with $C^l$ dimensions. The attention weight $\alpha_{i}^l$ is computed as:
	\begin{equation} \label{att_l}
	\alpha_{i}^l=\text{ReLU}((\bm{W}_{\alpha}^l)^{\top}(\bm{U}^l\bm{f}_i\lVert\bm{V}^l\bm{f}_i^l)),
	\end{equation}
	where $\bm{W}_{\alpha}^l$, $\bm{U}^l$, and $\bm{V}^l$ are defined similarly to those in \eqref{att_r}.
	
	\subsubsection{Global Contrast Graph}
	Similarly, we also consider incorporating global contrast inference in our graph model. A straightforward way to do so is to build a global context node with the global image feature. However, we argue that a single global node is too coarse and extracting only one global feature causes severe information loss. Instead, we construct a set of global context nodes by using non-overlapped average-pooling on $\bm{F}$ and obtain an $M\times M$ global context feature map. Hence, we construct $M^2$ global context nodes and connect all of them to each instance for inferring the global contrast, as shown in Figure~\ref{fig:graph}. We denote each global context node as $\mathbb{G}_i$ and its feature as $\bm{f}_i^g$, where $i\in\{1,2,\cdots,M^2\}$. Then, we can construct a global contrast graph as $\mathcal{G}^g=(\mathcal{V}^g,\mathcal{E}^g)$, where $\mathcal{V}^g=\{\{\mathbb{I}_i\}_{i=1}^{N},\{\mathbb{G}_j\}_{j=1}^{M^2}\}$. The edge set $\mathcal{E}^g$ has $M^2\times N$ edges, and each edge $e_{j,i}^g$ connects a global context node $\mathbb{G}_j$ to an instance node $\mathbb{I}_i$. For $\mathbb{I}_i$, its global contrast neighbor $\mathcal{N}_i^g$ is equal to all the global context nodes $\{\mathbb{G}_j\}$. Hence, its aggregator function has the formulation:
	\begin{equation} \label{agg_g}
	\bm{h}_{\mathcal{N}_i^g}=\sum_{j=1}^{M^2}\alpha_{ji}^g\bm{W}_a^g\bm{f}_j^g,
	\end{equation}
	where $\bm{W}_a^g\in{\mathbb{R}^{C^g\times 1024}}$, and 
	\begin{equation} \label{att_g}
	\alpha_{ji}^g=\frac{1}{M^2}\text{ReLU}((\bm{W}_{\alpha}^g)^{\top}(\bm{U}^g\bm{f}_i\lVert\bm{V}^g\bm{f}_j^g)).
	\end{equation}
	Here $\bm{W}_{\alpha}^g$, $\bm{U}^g$, and $\bm{V}^g$ have similar definitions to previous equations.
	
	\subsubsection{Incorporating Person Prior}
	Since we have found person to be one of the most salient object categories from previous works \cite{judd2009learning,borji2012boosting} and Figure~\ref{fig:category_score}, we incorporate the person prior in our proposed graph reasoning module. First, we construct a person classification head in parallel with the box head and the mask head in Mask R-CNN, as shown in Figure~\ref{network_fig}. Specifically, similar to the box head, we use two FC layers with 1024 channels to classify each instance $\mathbb{I}_i$ as belonging to the person class or not. Then, we treat the last FC feature as the person prior feature $\bm{f}_i^p$ and construct a person prior node $\mathbb{P}_i$ based on it. As such, we can build a person prior graph $\mathcal{G}^p=(\mathcal{V}^p,\mathcal{E}^p)$, as shown in Figure~\ref{fig:graph}. It is similar to the local contrast graph, \ie $\mathcal{V}^p=\{\mathbb{I}_i, \mathbb{P}_i\}_{i=1}^{N}$ and $\mathcal{E}^p=\{e_{i}^p\}_{i=1}^{N}$, where $e_{i}^p$ connects $\mathbb{P}_i$ to $\mathbb{I}_i$. For $\mathbb{I}_i$, the person prior neighbor $\mathcal{N}_i^p$ is equal to $\mathbb{P}_i$ and the aggregator function is defined as:
	\begin{equation} \label{agg_p}
	\bm{h}_{\mathcal{N}_i^p}=\alpha^p\bm{W}_a^p\bm{f}_i^p,
	\end{equation}
	where $\bm{W}_a^p\in{\mathbb{R}^{1024\times 1024}}$ transforms the original person prior feature to the graph learning space and $\alpha^p$ is a learnable parameter to weight the person prior information. Here we do not adopt the dynamic attention mechanism anymore since prior knowledge is usually integrated in a static way.
	
	\subsubsection{Graph Updating and Saliency Ranking Inference}
	After aggregating messages from the four graphs, we integrate them into an overall graph $\mathcal{G}$ and update each instance representation to obtain the graph output with a residual connection \cite{he2016resnet}:
	\begin{equation} \label{output}
	\bm{f}_i^{u}=\bm{f}_i+\bm{W}_{u}^r\bm{h}_{\mathcal{N}_i^r}+\bm{W}_{u}^l\bm{h}_{\mathcal{N}_i^l}+\bm{W}_{u}^g\bm{h}_{\mathcal{N}_i^g}+\bm{h}_{\mathcal{N}_i^p},
	\end{equation}
	where $\bm{W}_u^*\in{\mathbb{R}^{1024\times C^*}}$ projects the corresponding aggregated message back to the instance feature space. %In the graph model, $C^r$ is usually set to half of the original channel number, \ie 512.
	
	To stabilize the learning process and enrich the learned node interaction connections, we extend the overall graph $\mathcal{G}$ into $K$ parallel subgraphs, each of which independently learns node interactions. Finally, we concatenate the output of each subgraph as the final update signal:
	\begin{equation} \label{agg_multi}
	\bm{f}_i^{u}=\bm{f}_i+\overset{K}{\underset{k=1}{\Big\lVert}}(\bm{W}_{u}^{r,k}\bm{h}_{\mathcal{N}_i^r}^k+\bm{W}_{u}^{l,k}\bm{h}_{\mathcal{N}_i^l}^k+\bm{W}_{u}^{g,k}\bm{h}_{\mathcal{N}_i^g}^k+\bm{h}_{\mathcal{N}_i^p}^k).
	\end{equation}
	In this case, $C^*$ in each subgraph is usually set to $1024/K$ to reduce the computational costs and the size of $\bm{W}_u^{*,k}$ is changed to $\frac{1024}{K}\times \frac{1024}{K}$. For the person prior graph, the size of $\bm{W}_a^p$ is also changed to $\frac{1024}{K}\times 1024$.
	
	After obtaining the updated instance feature $\bm{f}_i^{u}$, we can directly use an FC layer with one channel to regress a saliency score $s_i$. Then, we can rank the scores of all segmented instances and obtain the rank order. Finally, we assign each rank order to the corresponding instance mask, obtaining the saliency ranking map, as shown in Figure~\ref{network_fig}.
	
	\subsection{Loss Function}
	In \cite{amirul2018rsd}, Islam \etal trained their RSDNet based on the pixel-wise Euclidean loss between the predicted saliency map and the GT. Siris \etal \cite{siris2020inferring} treated saliency ranking as a rank order classification problem, where they adopted a Softmax classifier with the cross entropy loss to categorize each instance into one of the five rank orders. In this paper, we concentrate on predicting rank orders for each image with varying salient instance numbers. Thus, we propose to train the saliency ranking branch using a ranking loss with respect to the GT rank order. Inspired by \cite{chen2016single}, we adopt a pair-wise ranking loss to encourage increasing saliency values for high-rank instances and decreasing saliency values for low-rank ones. Concretely, considering a training image with $N$ instances, the GT ranks are denoted as $\{r_1,r_2,\cdots,r_N\}$, where $r_i\in{\{1,2,\cdots,N\}}$, and smaller numbers indicate higher ranks. We extract all $C_N^2$ instance pairs for training. For a pair $q$, we rank its two instances according to the GT ranks, \ie we represent $q$ as $q=\{q_1,q_2 \}$ where $q_1,q_2\in{\{1,2,\cdots,N\}}$ and $r_{q_1}<r_{q_2}$. By denoting the predicted saliency scores of the two instances as $s_{q_1}$ and $s_{q_2}$, we have the ranking loss as:
	\begin{equation} \label{loss_org}
	L=\frac{1}{C_N^2}\sum_{q=1}^{C_N^2}log(1+exp(-s_{q_1}+s_{q_2})).
	\end{equation}
	
	However, such a loss treats all instance pairs equally and does not consider the specific GT rank orders. As such, it is harmful for optimizing the saliency scores of instances with very high or very low ranks. Hence, we add a dynamic loss weight $\beta$ to assign large weights for pairs with large rank differences and small weights for those with close ranks, thus explicitly optimizing instances with extreme ranks. Concretely, the ranking loss is improved as:
	\begin{equation} \label{loss_ours}
	L=\sum_{q=1}^{C_N^2}\beta_q log(1+exp((-s_{q_1}+s_{q_2}))),
	\end{equation}
	where $\beta_p$ is set according to the rank difference of $q_1$ and $q_2$, and normalized by:
	\begin{equation} \label{beta}
	\beta_q=\frac{(r_{q_1}-r_{q_2})^\gamma}{\sum_{o=1}^{C_N^2}(r_{o_1}-r_{o_2})^\gamma},
	\end{equation}
	where $\gamma>0$ and the larger it is, the larger the weights are for pairs with large rank differences.
	
	%------------------------------------------------------------------------
	\section{Experiments}
	\subsection{Implementation Details}
	To facilitate network training, we first pre-train our improved Mask R-CNN on the agnostic instance segmentation task, which only differentiates objects from non-objects in the box head. Specifically, we pre-train the improved Mask R-CNN on the MS-COCO \cite{lin2014coco} 2017 training split. Random horizontal flipping and the multi-scale training trick are used for data augmentation. Most training settings are the same as in \cite{he2017mask}, including the anchor settings, the GT box assignment scheme in RPN, and the non-maximum suppression (NMS) parameters, etc. We use the stochastic gradient descent (SGD) algorithm to train the network and set the batchsize and the maximum iteration step to 4 and 540,000, respectively. We first use the warm-up strategy to train the network with small learning rates in the first 1,000 steps and then train it with a normal learning rate which starts from 5e-3 and is divided by 10 at the 420,000th and the 500,000th steps, respectively.
	
	Then, we add the saliency ranking branch to the pre-trained improved Mask R-CNN and finetune the whole network on the training set of our proposed dataset. The person classification head is also trained from scratch in this stage. For simplicity, we use GT instance boxes to train the saliency ranking branch using the loss from Eq.~\eqref{loss_ours}. We also keep finetuning the improved Mask R-CNN using instance annotations. Compared with the pre-training stage, \rev{herein we use a fixed training size as $480\times 640$ and do not adopt any data augmentation techniques. We also change the batchsize and the iteration step to 1 and 200,000, respectively. The Adam \cite{kingma2014adam} algorithm is selected as the optimizer and the learning rate is set to $1e-5$.} For learning rate decay, we divide it by 10 at the 150,000th and the 180,000th steps, respectively.
	
	The whole model is implemented using the Pytorch \cite{paszke2017pytorch} framework. When testing, we directly resize each image to $480\times 640$ pixels as the network input. After obtaining salient instances from the Mask R-CNN, we filter out the output instances by NMS and also discard those having small ($<0.6$) saliency confidence scores obtained by the box head. Finally, we input the remaining instance boxes into the saliency ranking branch to obtain their rank order. Our dataset and code will be released.
	
	\subsection{Evaluation Metric}
	\begin{figure}[!t]
		\graphicspath{{Figures/SASOR/}}
		\centering
		\includegraphics[width=1\linewidth]{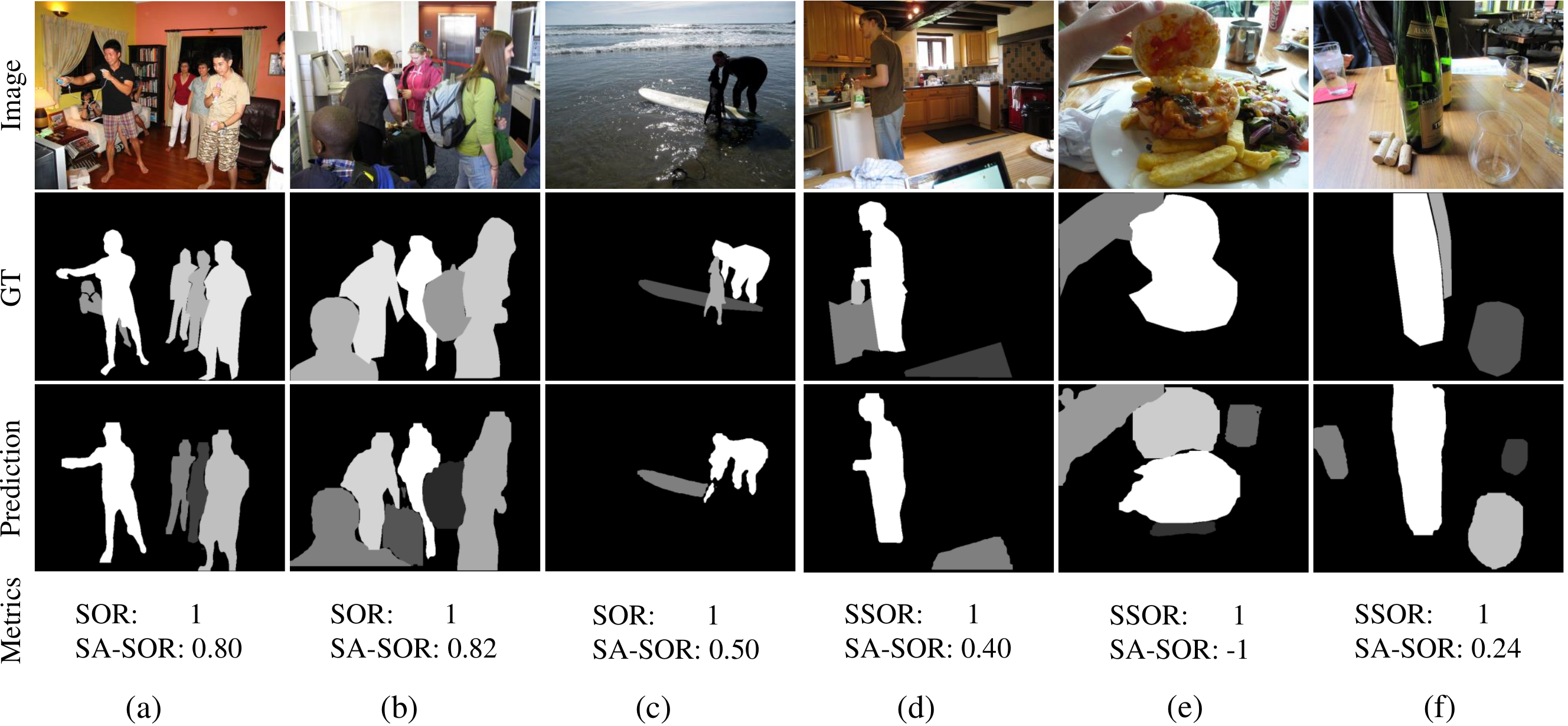}
		\caption{\textbf{Illustration of the limitations of the SOR metric \cite{amirul2018rsd} and the SSOR metric \cite{siris2020inferring}}. Both of them can not include the consideration of the segmentation quality into the evaluation satisfactorily.}
		\label{fig:metric_cmp}
		\vspace{-0.2cm}
	\end{figure}
	
	In \cite{amirul2018rsd}, the authors used GT instance masks to obtain saliency ranks and introduced the SOR metric to evaluate the salient instance ranking performance. SOR computes Spearman's rank-order correlation between the predicted rank order and the GT rank order of the salient instances in each image. Then, the SOR score is normalized to [0,1] and averaged over the whole dataset. However, their SOR metric assumes that the prediction has the same instances as the GT and only cares about rank orders. \rev{For example, as shown in Figure~\ref{fig:metric_cmp}(a), (b), (c), high SOR scores can be obtained as long as the detected salient instances having correct rank orders, despite the existence of missed or redundantly detected instances, or low-quality segmentation.} For the instance-level saliency ranking task, one should learn to simultaneously segment salient instances and rank their orders. Hence, the evaluation metric should also be sensitive to the segmentation \rev{quality}. Siris \etal \cite{siris2020inferring} proposed to match GT instance masks with the segmented ones by finding the segmented instance with the largest mask area in each GT mask, thus partially relieving the segmentation-unawareness problem of the original SOR metric. However, their naive matching method cannot guarantee strict one-to-one matching, and thus may introduce ambiguity. Furthermore, they directly ignore missed instances and redundantly segmented ones, preventing their evaluation method from being strictly sensitive to segmentation performance. \rev{We refer to Siris' SOR metric as SSOR and show some examples of their incorrect scores in Figure~\ref{fig:metric_cmp}(d), (e), and (f).}
	%加一部分讨论我们的metric和他们的metric，可以给出几个图示例
	
\begin{figure}[h]
		\graphicspath{{Figures/SASOR/}}
		\centering
		\includegraphics[width=1\linewidth]{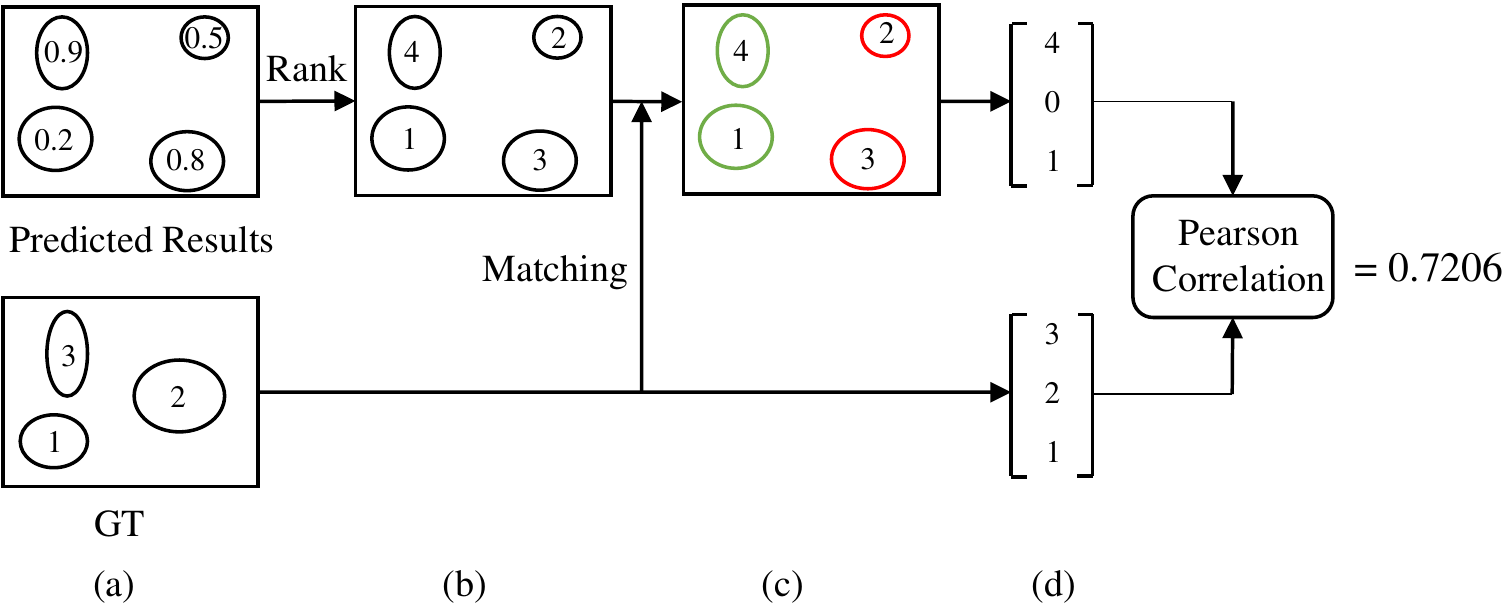}
		%\begin{overpic}[width=1\linewidth,grid,tics=2]{intro.pdf}
		%\begin{overpic}[width=1\linewidth]{SASOR.pdf}
		%\end{overpic}
		\caption{\textbf{Illustration of the computation process of the proposed SA-SOR metric.} We use circles to denote instance masks. We first rank the segmented instances based on their predicted saliency scores, as shown in columns (a) and (b). We use an ascending rank order for both predicted ranks and GT ranks, \ie larger rank values indicate higher degrees of saliency and 1 is the lowest rank. Then, we match the segmented instances with the GT, obtaining the matching results in column (c), 
%		where {\color{green}green} circles indicate matched instances and {\color{red}red} circles represent instances that cannot be matched with the GT. 
		where green circles indicate matched instances and red circles represent instances that cannot be matched with the GT.
		Next, we obtain the predicted rank order in column (d) by skipping the rank orders of the $3^{rd}$ and $2^{nd}$ segmented instances (which are redundant) and setting the rank of the $2^{nd}$ instance (which is missed in the segmentation result) as 0. Finally, we compute the Pearson correlation
			%between the predicted rank order and the GT one
			as the SA-SOR score.}
		\label{sasor_fig}
		\vspace{-0.3cm}
	\end{figure}

	To this end, we propose a segmentation-aware SOR (\textbf{SA-SOR}) metric to more strictly incorporate the influence of the alignment between segmented salient instances and the GT masks. To be specific, as shown in Figure~\ref{sasor_fig}, for the segmented instances of a test image, we first rank their predicted saliency scores and assign the ranks to each instance. To facilitate computation and understanding, we use an ascending rank order for both predicted ranks and the GT ranks, \ie larger rank values indicate higher degrees of saliency, and 1 is the lowest rank. Then, we match the segmented instance masks and GT masks with an IoU threshold of $t$ (which we set to 0.5 in this work). Here, we follow the matching strategy of computing the average precision (AP) metric in the instance segmentation task, \ie segmented masks are assigned to GT masks that satisfy an overlapping criterion according to a descending order of instance confidences. Finally, each GT instance can only be matched with at most one segmented instance and vice versa, where the IoU of two matched masks should be greater than or equal to $t$.
	
	Next, as shown in column (d) of Figure~\ref{sasor_fig}, we pick out the ranks of the matched instances and set the ranks of missed instances to 0, thus obtaining a predicted rank order. Then, we compute the SA-SOR score as the Pearson correlation between the predicted rank order and the GT, without using normalization as done in \cite{amirul2018rsd}. As such, redundantly segmented instances (false positives, shown as red circles in column (c) of Figure~\ref{sasor_fig}) will displace the predicted rank order and the missing instances can only have a rank of 0, which will both lower the SA-SOR score, making it strictly sensitive to the segmentation performance. As a result, the proposed SA-SOR score will encourage both accurate salient instance segmentation and correct ranking orders.
	
	%-------------------------------------------------------------------------
	\subsection{Ablation Study}\label{sec:ablation}
	In this section, we thoroughly evaluate the effectiveness of our different model settings on the test set of our proposed dataset.
	
	\subsubsection{Effectiveness on the Proposed Graph Models}
	
	\begin{table} [h]
		\begin{center}
			\caption{\textbf{Ablation study on the effectiveness of the four proposed graph reasoning models.} \rev{Baseline I means we directly regress the saliency value of each instance based on its instance feature $\bm{f}_i$ while baseline II means we use all the four kinds of features to regress the instance saliency.} \fst{Bold} indicates the best performance.}
			\label{tab:graph_cmp}
			\small
			\begin{tabular}{@{}L{1.6cm}|L{2cm}|L{2cm}|C{1.2cm}}
				\toprule
				Settings & Features  & Graphs  & SA-SOR
				\\ \midrule
				I (Baseline)         &$\bm{f}_i$                                        &           &0.5147
				\\
				II (Baseline)        &$\bm{f}_i$+$\bm{f}_i^l$+$\bm{f}^g$+$\bm{f}_i^p$   &           &0.5419
				\\ \midrule
				III                  &$\bm{f}_i$                                        &$\mathcal{G}^r$           &0.5339 
				\\
				IV                   &$\bm{f}_i$+$\bm{f}_i^l$                          &$\mathcal{G}^r$+$\mathcal{G}^l$         &0.5487
				\\
				V                    &$\bm{f}_i$+$\bm{f}_i^l$+$\bm{f}^g$                &$\mathcal{G}^r$+$\mathcal{G}^l$+$\mathcal{G}^g$       &0.5568 
				\\
				VI                   &$\bm{f}_i$+$\bm{f}_i^l$+$\bm{f}^g$+$\bm{f}_i^p$   &$\mathcal{G}^r$+$\mathcal{G}^l$+$\mathcal{G}^g$+$\mathcal{G}^p$    &\fst{0.5616}
				\\
				\bottomrule
			\end{tabular}
			\vspace{-0.2cm}
		\end{center}{}
	\end{table}
	
	\begin{figure*}[!t]
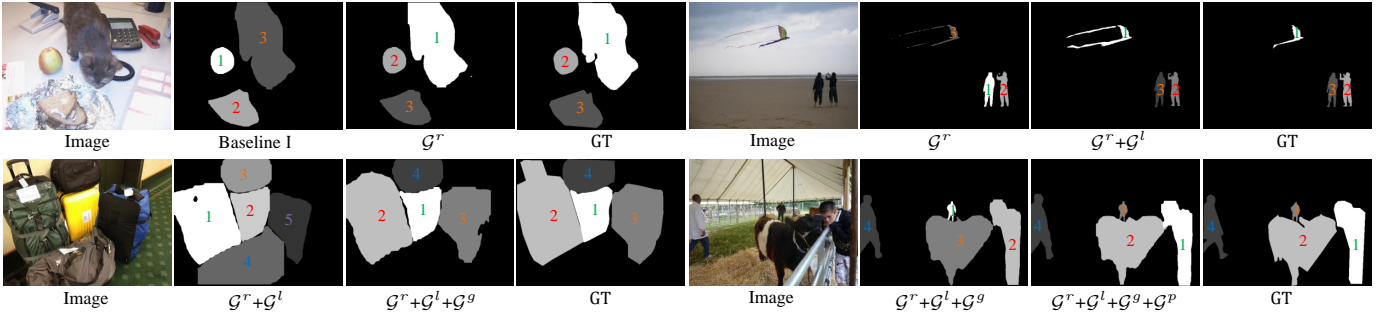

		\graphicspath{{Figures/ablation/}}
		\centering
		\begin{overpic}[width=1\linewidth]{qualitative.pdf}
			%		\begin{overpic}[width=1\linewidth,grid,tics=1]{qualitative.pdf}
			\put(30.6,11.7){\scriptsize $\mathcal{G}^r$}
			\put(67.6,11.7){\scriptsize $\mathcal{G}^r$}
			\put(79.5,11.7){\scriptsize $\mathcal{G}^r$+$\mathcal{G}^l$}
			\put(16.5,0){\scriptsize $\mathcal{G}^r$+$\mathcal{G}^l$}
			\put(28,0){\scriptsize $\mathcal{G}^r$+$\mathcal{G}^l$+$\mathcal{G}^g$}
			\put(65.2,0){\scriptsize $\mathcal{G}^r$+$\mathcal{G}^l$+$\mathcal{G}^g$}
			\put(76.6,0){\scriptsize $\mathcal{G}^r$+$\mathcal{G}^l$+$\mathcal{G}^g$+$\mathcal{G}^p$}
		\end{overpic}
		\caption{\textbf{Qualitative comparison among models using different graphs.} We show the comparison among five models in Table~\ref{tab:graph_cmp}, including the baseline model I, the model $\mathcal{G}^r$ in row III in Table~\ref{tab:graph_cmp}, the model $\mathcal{G}^r$+$\mathcal{G}^l$ in row IV, the model $\mathcal{G}^r$+$\mathcal{G}^l$+$\mathcal{G}^g$ in row V, and the model $\mathcal{G}^r$+$\mathcal{G}^l$+$\mathcal{G}^g$+$\mathcal{G}^p$ in row VI.}
		\label{fig:abl_qual}
		\vspace{-0.3cm}
	\end{figure*}
	
	In our graph reasoning model, we build four graphs to incorporate instance interaction relations, local and global contrast inference, and the person prior knowledge. Hence, in this part, we evaluate the effectiveness of the four graphs. \rev{We start from the baseline models in which no graph network is used. As shown in Table~\ref{tab:graph_cmp}, our first baseline model directly regresses the saliency score for each instance based on its instance feature $\bm{f}_i$ using two FC layers with 1024 and 1 channel, respectively. In our second baseline model, we leverages all four kinds of features, \ie the instance feature $\bm{f}_i$, the local context feature $\bm{f}_i^l$, the global context feature $\bm{f}^g$, and the person prior feature $\bm{f}_i^p$. Specifically, we first fuse the $M\times M$ global context feature map $\bm{f}^g$ into a single feature vector via an FC layer with 1024 channels and then combine it with the other three features via concatenation. The same two FC layers are adopted to regress the saliency value.
		Here the original ranking loss \eqref{loss_org} is used for training the saliency ranking branch. The comparison results of gradually adding the four graphs are shown from row III to row VI. We observe that using $\mathcal{G}^r$ largely improves the SA-SOR score compared with the baseline model I, indicating the importance of modeling the interaction relations among instances. Further adding $\mathcal{G}^l$, $\mathcal{G}^g$, and $\mathcal{G}^p$ can gradually improve the model performance, which demonstrates the effectiveness of our introduced local context, global context, and the person prior. By comparing baseline model II with model VI which use the same features, we can conclude that our proposed graph reasoning networks largely improve the model performance. As a result, our final model achieves significant model improvements compared with the baseline model I.}
	
	\rev{We also show some qualitative comparison results in Figure~\ref{fig:abl_qual} to better understand the effectiveness of our proposed graph models. At the top left, we first show the comparison between the baseline model I and the model with only $\mathcal{G}^r$, \ie the model III in Table~\ref{tab:graph_cmp}. We can see that with the help of modeling instance interaction and competition relations, the correct ranking order can be obtained for the cat. At the top right, if we only use $\mathcal{G}^r$, the kite will be treated as the least salient since usually persons are more salient. However, when using $\mathcal{G}^l$, our model can find that the kite has stronger local contrast, thus predicting correct ranks. At the bottom left, the perceived global contrast of the suitcases in $\mathcal{G}^g$ helps to rank them correctly, especially for the yellow one. Finally, at the bottom right, the person prior knowledge introduced by $\mathcal{G}^p$ not only helps infer the relative degree of saliency between the cow and the persons, but also helps judge the correct rank orders among persons, which indicates that our introduced person prior probably learns fine-grained person attributes, such as poses and expressions, for detailed saliency ranking inference.}

	\subsubsection{Influence of the Subgraph Number}
	
	\begin{table} [!t]
		\begin{center}
			\caption{\textbf{Influence of the subgraph number $K$ on the interaction relation graph $\mathcal{G}^r$.} \fst{Bold} indicates the best performance.}
			\label{tab:subgraph_cmp}
			\small
			\begin{tabular}{@{}L{1.1cm}|C{2.1cm}|C{1.4cm}}
				\toprule
				Settings & Subgraphs ($K$)  & SA-SOR
				\\\midrule
				I                    &1           &0.5223
				\\
				II                   &4           &0.5160
				\\
				III                  &8           &\fst{0.5339}
				\\
				IV                   &16          &0.5164
				\\
				\bottomrule
			\end{tabular}
			\vspace{-0.3cm}
		\end{center}{}
	\end{table}
	
	Since we use $K$ parallel subgraphs for our proposed graph models to stabilize the learning process and enrich the learned graph connections, it is also very important to explore how many subgraphs should be used. To this end, we conduct experiments using different numbers of subgraphs. For simplicity, we test the influence of $K$ only on the interaction relation graph $\mathcal{G}^r$, and the results are shown in Table~\ref{tab:subgraph_cmp}. We can see that using different numbers of subgraphs leads to different SA-SOR scores, and setting $K=8$ achieves the best performance. Hence, we use eight subgraphs as default.
	
	\subsubsection{Influence of the Number of Global Context Nodes}
	
	In our global contrast graph $\mathcal{G}^g$, we construct $M \times M$ global context nodes using adaptive average-pooling. Hence, we also verify how many global context nodes are the best for our global contrast graph model. We test four different values for $M$, \ie 3, 7, 10, and 14, and report the model performance in Table~\ref{tab:M_cmp}. The results show that, generally, more global context nodes bring better model performance, which means fine-grained global contexts are better for global contrast inference than a coarse global representation. However, the model performance becomes saturated when $M\geqslant 7$. Hence, we select $M=7$ as our model setting.
	
	\begin{table} [h]
		\begin{center}
			\caption{\textbf{Ablation study on how many global context nodes should be used in $\mathcal{G}^g$.} \fst{Bold} indicates the best performance.}
			\label{tab:M_cmp}
			\small
			\begin{tabular}{@{}L{1.1cm}|C{2.1cm}|C{1.4cm}}
				\toprule
				Settings & $M$  & SA-SOR
				\\\midrule
				I                    &3           &0.5455
				\\
				II                   &7           &\fst{0.5568}
				\\
				III                  &10          &0.5566
				\\
				IV                   &14          &\fst{0.5568}
				\\
				\bottomrule
			\end{tabular}
			\vspace{-0.2cm}
		\end{center}{}
	\end{table}
	
	\begin{table} [!t]
		\begin{center}
			\caption{\textbf{Ablation study on the effectiveness of the proposed ranking loss.} We first conduct a comparison against the Softmax with cross entropy loss (CEL) used in \cite{siris2020inferring}. Then, we vary the value of $\gamma$ in \eqref{beta} to verify the usefulness of our proposed ranking loss (RL). Setting $\gamma=0$ downgrades to the original ranking loss in \eqref{loss_org}. \fst{Bold} indicates the best performance.}
			%\vspace{-0.2cm}
			\label{tab:loss_ablation}
			\small
			\begin{tabular}{@{}L{1.1cm}|L{2.6cm}|C{1.4cm}}
				\toprule
				Settings & Loss  & SA-SOR
				\\ \midrule
				I                     &Softmax+CEL \cite{siris2020inferring}                   &0.5271
				\\ \midrule
				II                    &RL ($\gamma=0$) \eqref{loss_org}       &0.5616
				\\
				III                   &RL ($\gamma=0.25$)      &0.5521
				\\
				IV                    &RL ($\gamma=0.5$)       &0.5602
				\\
				V                     &RL ($\gamma=0.75$)      &0.5626
				\\
				VI                    &RL ($\gamma=1$)         &\fst{0.5647}
				\\
				VII                    &RL ($\gamma=1.25$)      &0.5476
				\\
				VIII                    &RL ($\gamma=1.5$)       &0.5345
				\\
				\bottomrule
			\end{tabular}
			\vspace{-0.3cm}
		\end{center}{}
	\end{table}
	
	\subsubsection{Effectiveness on the Proposed Ranking Loss}
	
	We first evaluate the effectiveness of the proposed ranking loss by comparing it against the traditional Softmax with cross-entropy loss used in \cite{siris2020inferring}. Specifically, based on our best-performing model with all four proposed graphs, we mimic the training protocol in \cite{siris2020inferring} by employing an eight-channel Softmax classifier with the cross entropy loss to classify the instance ranks in each image. Then, we vary the value of $\gamma$ in \eqref{beta} to verify the usefulness of our proposed dynamic loss weight $\beta$ in \eqref{loss_ours} (note that setting $\gamma=0$ degrades our improved loss back to the original ranking loss in \eqref{loss_org}). The results are given in Table~\ref{tab:loss_ablation}. We can first see that using the traditional Softmax classifier leads to a significant performance drop, demonstrating the superiority of modeling the saliency rank task as a ranking problem instead of a classification task. Then, by varying the values of $\gamma$, we also find that our proposed dynamic loss weight can further improve the model performance. Concretely, setting $\gamma$ to 1 leads to the best performance, while setting it larger than 1 will largely downgrade the model performance, which may be caused by the extreme loss weights when using too large $\gamma$.
	
	\subsection{Comparison with State-of-the-Art Methods}
	
	\begin{table*} [h]
		\begin{center}
			\caption{\textbf{Quantitative comparison of our proposed model with RSDNet \cite{amirul2018rsd} and ASSR \cite{siris2020inferring} on three datasets.} For fair and comprehensive comparisons, we train and test all three models on all three datasets. In addition to the proposed SA-SOR metric, \rev{the original SOR or the SSOR} metric and MAE are also adopted for a comprehensive evaluation. \rev{Specifically, since RSDNet can not predict salient instances, we adopt the SOR metric for it. For ASSR and our model, we use the SSOR metric. FPS is considered to evaluate the model efficiency as well.} Please note that on the PASCAL-S dataset, we only test on images with more than one rank. Hence, the results of RSDNet may be different from those in the original paper \cite{amirul2018rsd}. \fst{Bold} indicates the best performance.}
			\label{tab:ranking_saliency_cmp}
			%\small
			\footnotesize
			\begin{tabular}{@{}R{1.5cm}|C{0.6cm}|C{1.3cm}C{1.3cm}C{0.65cm}|C{1.3cm}C{1.3cm}C{0.65cm}|C{1.3cm}C{1.3cm}C{0.65cm}@{}}
				\toprule
				\multirow{2}{*}{Models} &\multirow{2}{*}{FPS} &\multicolumn{3}{c|}{PASCAL-S \cite{li2014secrets}} &\multicolumn{3}{c|}{Siris' Dataset \cite{siris2020inferring}} &\multicolumn{3}{c}{Ours Dataset}
				\\  \cmidrule{3-11}
				&&SA-SOR &SOR/SSOR &MAE &SA-SOR &SOR/SSOR &MAE &SA-SOR &SOR/SSOR &MAE \\ \midrule
				RSDNet \cite{amirul2018rsd} &
				4.6 &0.695  & 0.852  & 0.241  & 0.499  & 0.717  & 0.158  & 0.460  & 0.735  & 0.129
				\\
				ASSR \cite{siris2020inferring} &
				2.2 &0.644  & 0.847  & 0.112  & 0.667  & 0.792  & \fst{0.101}  & 0.388  & 0.714  & 0.125
				\\
				Ours&
				\fst{15.4} &\fst{0.739}  & \fst{0.869}  & \fst{0.106}  & \fst{0.709}  & \fst{0.811}  & 0.105  & \fst{0.565}  & \fst{0.806}  & \fst{0.085}
				\\
				\bottomrule
			\end{tabular}
			\vspace{-0.2cm}
		\end{center}{}
	\end{table*}
	
	\begin{figure*}[!t]
		\graphicspath{{Figures/qualitative/}}
		\centering
		\includegraphics[width=1\linewidth]{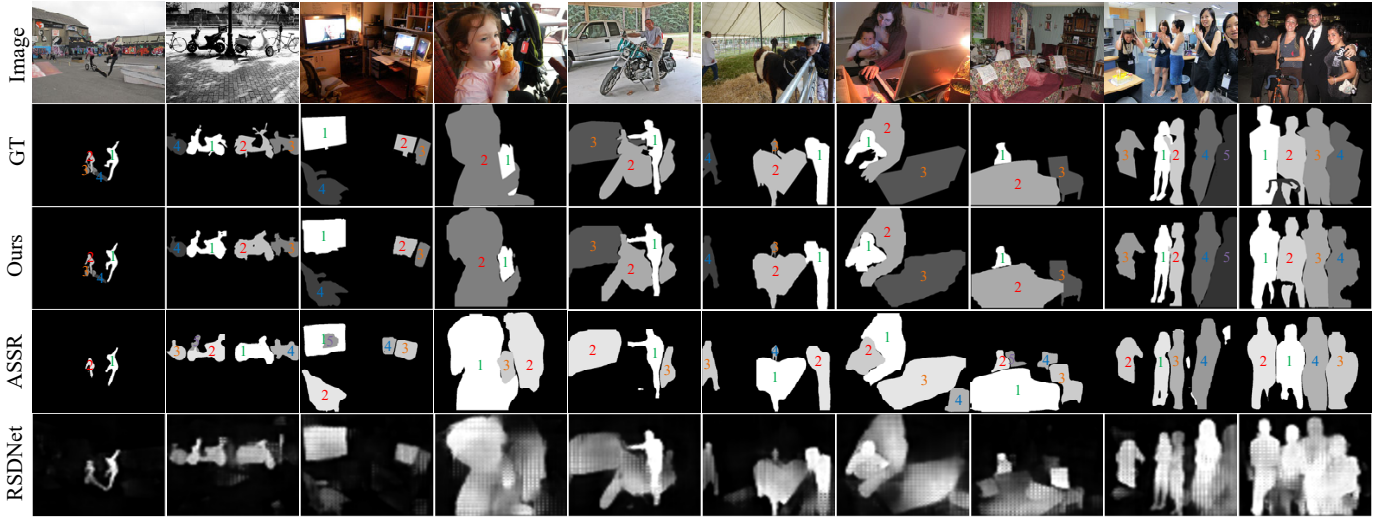}
		%\begin{overpic}[width=1\linewidth,grid,tics=2]{intro.pdf}
		%\begin{overpic}[width=1\linewidth]{SASOR.pdf}
		%\end{overpic}
		\caption{\textbf{Visual comparison of the saliency maps of our proposed model, RSDNet \cite{amirul2018rsd}, and ASSR \cite{siris2020inferring}.} Our model can segment salient instances and rank their saliency orders well, obtaining saliency maps very similar to the GT.
		}
		\label{vis_cmp_fig}
		\vspace{-0.3cm}
	\end{figure*}
	
	To verify the effectiveness of our proposed model on the saliency ranking prediction problem, in this section we compare it with other state-of-the-art saliency models, including saliency ranking models, SOD models, and EFP models.
	
	\subsubsection{Comparison with Saliency Ranking Methods}
	We compare our proposed model with two other state-of-the-art saliency ranking methods, \ie RSDNet \cite{amirul2018rsd} (we use their best-performing RSDNet-R model) and ASSR \cite{siris2020inferring}. For more comprehensive evaluations, we compare three models on all three datasets, \ie the PASCAL-S dataset, Siris' dataset, and our proposed dataset. Since the original three models are all separately trained on the three datasets, we further train each method on the other two datasets for a fair evaluation. Specifically, 
	\begin{compactitem}
		\item When training RSDNet on Siris' and our dataset, we directly change its output channel number to the maximum rank in each dataset, \ie 5 and 8, respectively, to regress the saliency values for each pixel.
		\item When training the ASSR model on the PASCAL-S dataset and our dataset, we also change its output channel number to the maximum rank in each dataset, \ie 12 and 8, respectively.
		\item Our proposed model can be directly trained on the PASCAL-S dataset and Siris' dataset. However, since the Siris' dataset only provides annotations for up to the five most salient instances in the GT saliency map, and the other two models can both benefit from fixed output channel numbers, we only select at most five instances in our prediction results for fair comparisons.
		\item For the PASCAL-S dataset, since it only includes 275 images with more than one rank to train ranking models, we conduct finetuning for both our model and the ASSR network.
	\end{compactitem}
	
	As for evaluation metrics, besides SA-SOR, we also adopt \rev{the original SOR or the SSOR} metric to ensure fair comparisons with \cite{amirul2018rsd,siris2020inferring}. \rev{Specifically, since RSDNet can not predict salient instances, we adopt the SOR metric for it. For ASSR and our model, we use the SSOR metric.} 
	%	When computing the SOR score for our prediction results, we follow \cite{siris2020inferring} and match each GT instance mask with our saliency map, assigning the predicted rank with the largest overlapped mask area to it. 
	As for computing the SA-SOR score for RSDNet, we propose to use the segmented instances of our improved Mask R-CNN as its instance masks and assign the averaged saliency value within each predicted mask as the saliency score of that instance. \rev{By following \cite{siris2020inferring}, the mean absolute error (MAE) metric, which directly considers the per-pixel saliency value difference between the predicted saliency map and the ground truth, is also adopted for a comprehensive evaluation. Furthermore, to evaluate the model efficiency in potential practical applications, we include the inference speed in terms of FPS in the comparison as well.}
	
	\begin{table*} [t]
		\begin{center}
			\caption{\textbf{Quantitative comparison of our proposed model with ten state-of-the-art SOD methods and nine EFP models.} We conduct the comparison on our proposed dataset in terms of the proposed SA-SOR metric \rev{and FPS}. \fst{Bold} indicates the best performance. \rev{Please note that all SOD models are re-trained using the stacked representation of ground truth \cite{amirul2018rsd} to learn relative saliency.}}
			\label{tab:sod_efp_cmp}
			%\small
			\footnotesize
			\begin{tabular}{@{}L{1.35cm}|C{1.5cm}C{1.2cm}C{1.2cm}C{1.2cm}C{1.2cm}C{1.2cm}C{1.2cm}C{1.2cm}C{1.2cm}C{1.2cm}@{}}
				\toprule
				& \multicolumn{10}{c}{SOD Models}
				\\ \cmidrule{2-11}
				& DHS &BMP &DGRL &PiCA &DSS &PoolNet &AFNet &CPD &MINet & ITSD
				\\
				&\cite{liu2016dhsnet} &\cite{zhang2018bmp} &\cite{wang2018dgrl} &\cite{liu2018picanet} &\cite{hou2019dssPAMI} &\cite{liu2019poolnet} &\cite{feng2019afnet} &\cite{wu2019cpd} &\cite{pang2020minet} &\cite{zhou2020itsd}
				\\ \midrule
				SA-SOR &
				%0.330 &     0.325 &     0.319 &     0.320 &     0.326&     0.312 &     0.346 &    0.305 & 0.318 & 0.300
				0.437 &     0.428 &     0.400 &     0.465 &     0.404&     0.462 &     0.442 &    0.358 & 0.325 & 0.468
				\\
				FPS &
				\fst{16.4} &     13.4 &     4.1 &     4.7 &     14.2&     13.8 &     15.4 &    15.5 & 12.8 & 14.9
				\\ \midrule
				& \multicolumn{9}{c|}{EFP Models} & \multirow{3}{*}{Ours}
				\\ \cmidrule{2-10}
				& DeepGaze II &SalGAN &CASNet &SAM &DVA &DSCLRCN &GazeGAN &UNISAL &\multicolumn{1}{c|}{MSI-Net} &
				\\
				&\cite{kummerer2017understanding} &\cite{pan2017salgan} &\cite{fan2018emotional} &\cite{cornia2018predicting} &\cite{wang2017dva} &\cite{liu2018dsclrcn} &\cite{che2019gaze} &\cite{droste2020unified} &\multicolumn{1}{c|}{\cite{kroner2020contextual}} &
				\\ \midrule
				SA-SOR
				&  0.487	&  0.509	&  0.508	&  0.555	&  0.544	&  0.552	&  0.501	&  0.544	&  \multicolumn{1}{c|}{0.558}   &  \fst{0.565}
				\\
				FPS &
				12.0 &     0.5 &     10.7 &     0.4 &     14.4&     7.9 &     15.5 &    14.0 & 12.0 & 15.4
				\\
				\bottomrule
			\end{tabular}
			\vspace{-0.3cm}
		\end{center}{}
	\end{table*}
	
	Table~\ref{tab:ranking_saliency_cmp} shows the comparison results. We observe that our model generally outperforms the other two methods by a large margin on all three datasets. Specifically, on the PASCAL-S dataset and the Siris' dataset, our model achieves much higher SA-SOR scores and performs slightly better in terms of other metrics. The only exception is that on Siris's dataset, our model obtains slightly higher MAE than the ASSR model. On our dataset, our model significantly outperforms the other two algorithms in terms of all metrics. \rev{Furthermore, our model is much faster than the other two methods in terms of FPS.} These results clearly demonstrate the effectiveness of our proposed saliency ranking model and its superiority \rev{and efficiency} for practical usage.
	
	We also show visual comparisons of the predicted saliency maps for the three models in Figure~\ref{vis_cmp_fig}. We can see that our model can both accurately segment salient instances and correctly rank their saliency orders, thus generating saliency maps similar to the GT. Moreover, it can handle various challenging scenarios, such as cluttered backgrounds and images with many instances ($>3$).
	%It can also handle ``stuff'' classes after finetuned on the PASCAL-S dataset (see the penultimate column). 
	In contrast, RSDNet and ASSR often rank instances incorrectly in challenging visual scenes.
	
	\subsubsection{Comparison with Salient Object Detection and Eye Fixation Prediction Methods}
	
	For a more comprehensive evaluation, we follow \cite{amirul2018rsd,siris2020inferring} and also compare state-of-the-art SOD models, including ITSD \cite{zhou2020itsd}, MINet \cite{pang2020minet}, CPD \cite{wu2019cpd}, AFNet \cite{feng2019afnet}, PoolNet \cite{liu2019poolnet}, DSS \cite{hou2019dssPAMI}, PiCA \cite{liu2018picanet}, DGRL \cite{wang2018dgrl}, BMP \cite{zhang2018bmp}, and DHS \cite{liu2016dhsnet}. Furthermore, since relative saliency ranking is closely related to visual attention and \cite{kalash2019rsd,siris2020inferring} and our work all construct datasets from the saliency maps or fixation data of the SALICON \cite{jiang2015salicon} dataset, we also compare with nine state-of-the-art EFP models, including DeepGaze II \cite{kummerer2017understanding}, SalGAN \cite{pan2017salgan}, CASNet \cite{fan2018emotional}, SAM \cite{cornia2018predicting}, DVA \cite{wang2017dva}, DSCLRCN \cite{liu2018dsclrcn}, GazeGAN \cite{che2019gaze}, UNISAL \cite{droste2020unified}, and MSI-Net \cite{kroner2020contextual}. Considering that our problem setting requires each model to have the ability to simultaneously segment salient instances and rank their relative saliency, we use our improved Mask R-CNN as an off-the-shelf salient instance detector for these models and focus on comparing their saliency ranking performance. \rev{Specifically, for the SOD models, since they can not learn the concept of relative saliency explicitly, we follow \cite{amirul2018rsd} to re-train them using the stacked representation of ground truth. Next, we average their predicted saliency values within each instance mask as the saliency scores of that instance. For the EFP models, considering they can learn the relative saliency concept and following the way we construct our dataset, we use their original implementations to generate saliency maps and take the maximum saliency value within each instance mask as the instance saliency score.} Then, we rank all the segmented instances for each image and report the SA-SOR scores on our proposed dataset. \rev{Again, FPS is considered to evaluate their model efficiency. However, please note that we measure the speed of both SOD and EFP models with considering the time costs of our improved Mask R-CNN salient instance detector for fair comparisons.}
	
	Table~\ref{tab:sod_efp_cmp} shows the comparison results with both SOD and EFP models. We can see that state-of-the-art SOD models perform worse on the relative saliency ranking task, which is reasonable since they were originally designed for binary saliency detection. On the other hand, EFP models achieve promising results and several of them obtain only slightly worse results than our method. However, please note that these models are usually trained on the whole training set of SALICON, which has much more images than our training set. Our model still outperforms them despite using fewer images for training the saliency ranking branch. \rev{In terms of the model efficiency, although our model can not achieve the highest FPS, it is one of the fastest methods and is only slightly slower than the fastest one, \ie DHSNet \cite{liu2016dhsnet}.}
	
	%In Figure~\ref{fig:sod_efp_vis_cmp}, we show some qualitative comparisons between our model and state-of-the-art SOD and EFP images. We can see that 
	
	\subsection{Analysis and Discussion}
	
\begin{figure}[t]
		\graphicspath{{Figures/error_analy/}}
		\centering
		\includegraphics[width=1\linewidth]{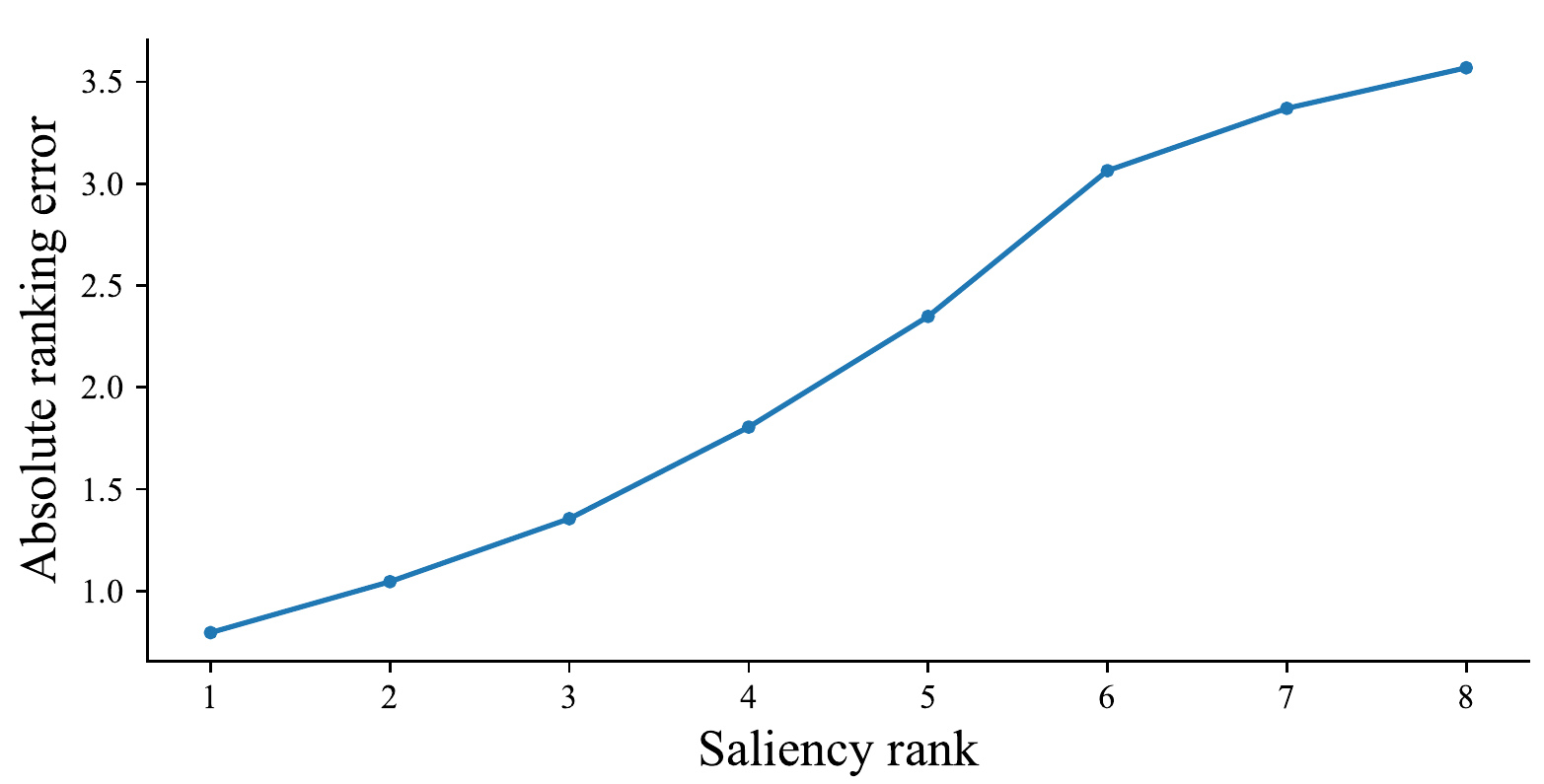}
		\caption{\textbf{Rank prediction error analysis.} We show the absolute ranking error for different saliency ranks.}
		\label{fig:error_analy}
		\vspace{-0.3cm}
	\end{figure}

	To analyze the ranking difficulty of relative saliency, we compute the absolute ranking error for each GT saliency rank on the test set of our dataset. Specifically, we use an \rev{descending order} to denote the predicted and GT saliency ranks and then compute the average absolute error between them for each rank order. The results are shown in Figure~\ref{fig:error_analy}. We find that the less salient the object is, the more difficult its rank order can be correctly predicted. For very low saliency ranks ($\geqslant 6$), our model encounters very large prediction errors ($\geqslant 3$). This is because less salient objects have subtle saliency differences, hence being difficult to handle. A possible solution is that we can use local feature maps to represent each instance node, thus facilitating mining these subtle differences. This observation also greatly encourages future works to improve the model capability in differentiating fine-grained saliency difference for less-salient objects.
	
	\subsection{Application: Adaptive Image Retargeting}
	Image retargeting aims at changing the aspect ratio or size of an image while preserving its important visual content. It is a typical application of saliency detection since saliency can be used as a useful cue to tell which regions are important and should be preserved, and which are negligible and can be removed. In \cite{setlur2007retargeting} and \cite{ren2009image}, Itti's saliency detection model \cite{itti1998model} is utilized along with face detectors to generate so-called energy maps, based on which seaming carving \cite{avidan2007seam} or warping can be applied to resize the input image in a content-aware way. Some more accurate saliency detection models, such as \cite{harel2007graph} and \cite{yan2013hierarchical}, were also adopted to obtain higher-quality energy maps in \cite{chen2015improved}, \cite{achanta2009saliency}. 
	%Subsequently, the depth information is also taken into consideration to assist saliency for higher quality energy map \cite{shafieyan2017image}, \cite{chen2015improved}.
	Recently, CNN-based approaches \cite{mastan2020dcil,tan2019cycle,cho2017weakly} were introduced into this task to produce more plausible target images by generating task-specific saliency maps internally.
	%However, the lack of a large-scale image retargeting dataset impedes the training of these deep models.
	These methods work well for images with a single salient object. However, for images with multiple foreground objects, things become different. If SOD saliency maps are adopted, their binary saliency modeling scheme makes it difficult to separate objects with different importance. Most previous methods adopt EFP saliency maps, which have the relative saliency concept but cannot detect object instances. Consequently, these methods may remove part of a salient instance and preserve other parts, making it incomplete. CNN-based deep retargeting models also have similar issues.
	
	\begin{figure*}[!t]
		\graphicspath{{Figures/seam_carving/}}
		\centering
		\includegraphics[width=1\linewidth]{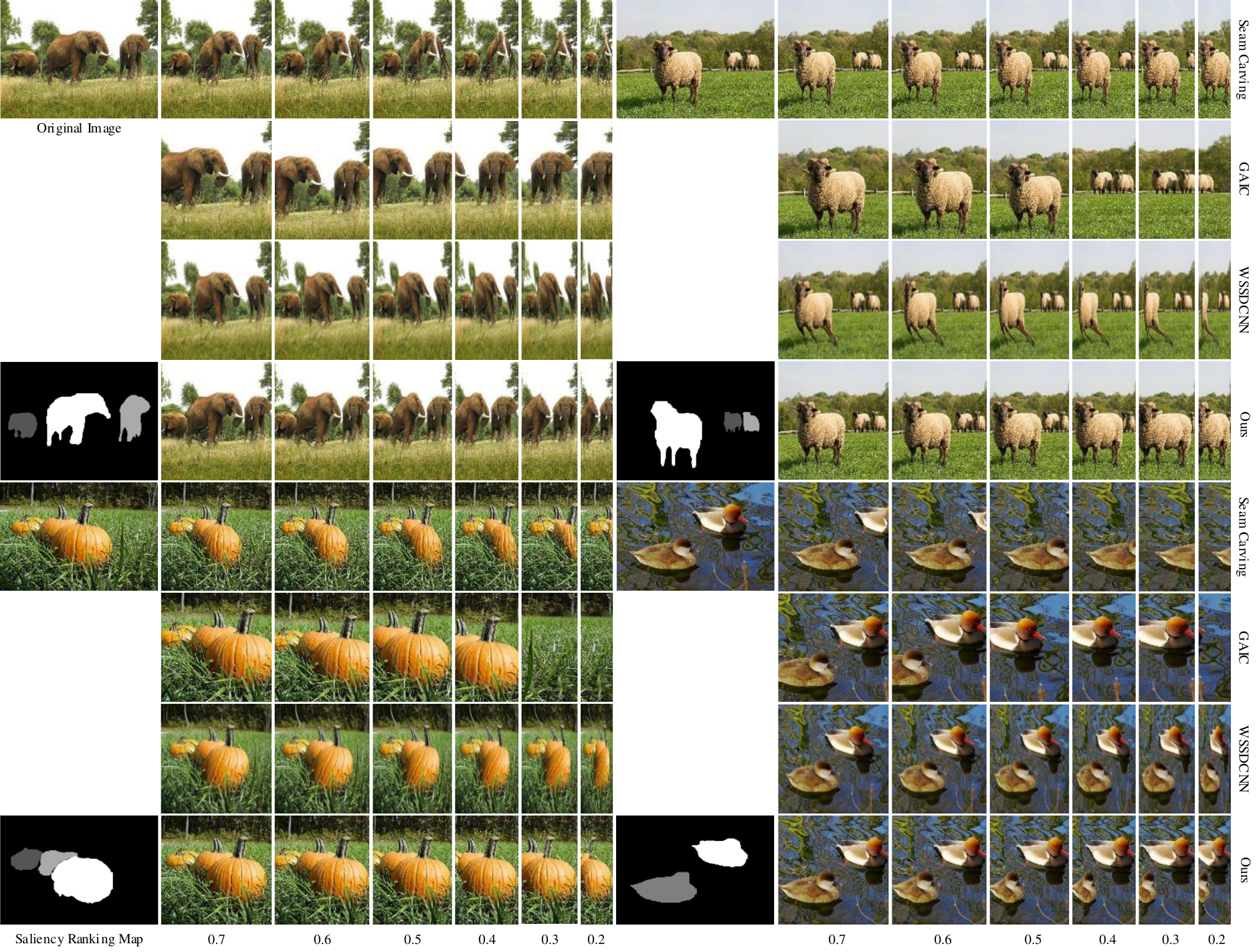}
		\caption{\textbf{Comparison of adaptive image retargeting results.} We show four groups of image retargeting results when progressively shrinking the width of each image from 0.7 times to 0.2 times of the original width. We include the original seam carving \cite{avidan2007seam} algorithm, GAIC \cite{zeng2019reliable}, and WSSDCNN \cite{cho2017weakly} for comparison. ``Ours" is a combination of our saliency ranking maps with seam carving. We also show our original saliency ranking map for each image.}
		\label{fig:seam_carving}
		\vspace{-0.3cm}
	\end{figure*}
	
	Considering an image with multiple salient instances, we can first remove background regions and shrink the objects with lower degrees of saliency when encountering small output size reduction. When facing large output size reduction, we can directly remove the objects with less importance and keep the most important ones. By treating different objects separately according to their degrees of saliency, more suitable and flexible retargeting results can be obtained for different target scales. We refer to this image retargeting scheme as adaptive image retargeting. Our proposed instance-level relative saliency ranking model ideally meets the requirement of this task.
	Specifically, we first run our model on each input image and then apply its saliency ranking map into the classical seam carving image retargeting algorithm \cite{avidan2007seam} by simply multiplying it with the original energy map. In this way, we leverage saliency ranking to modulate the priority of the regions to be preserved or removed.
	
	In Figure~\ref{fig:seam_carving}, we show some image retargeting results when progressively shrinking the width of each image. We compare the combination of our model and seam carving with the original seam carving \cite{avidan2007seam} algorithm, and two recently published methods, \ie the weakly- and self-supervised deep
	convolutional neural network (WSSDCNN) \cite{cho2017weakly} and grid anchor
	based image cropping (GAIC) \cite{zeng2019reliable}. We observe that the three compared models tend to generate severe distortions for the most important objects under small scale reduction, or even remove them entirely and preserve less important objects under large scale reduction. In contrast, combining our proposed ranking model with seam carving can lead to much more plausible retargeting results. The most salient objects are well preserved with less deformation, while less salient objects can be deformed first or removed under very large-scale reduction.

	\section{Conclusion}
	Despite its practical usage, instance-level relative saliency ranking detection has not been thoroughly investigated. In this paper, we present the first end-to-end solution for this task. Specifically, Mask R-CNN is first improved and used to segment salient instances. An additional graph neural network-based branch is added for saliency ranking inference. We investigate several aspects of the model effectiveness and identify several important considerations, for this new task, such as different saliency cues, model designs, and losses. For better model training and evaluation, we also propose a new dataset and an evaluation metric to benefit future research. Finally, experimental results demonstrate the effectiveness of our proposed model, including an example of its practical usage. We also analyze the model limitation and possible future works.

	% if have a single appendix:
	%\appendix[Proof of the Zonklar Equations]
	% or
	%\appendix  % for no appendix heading
	% do not use \section anymore after \appendix, only \section*
	% is possibly needed
	
	% use appendices with more than one appendix
	% then use \section to start each appendix
	% you must declare a \section before using any
	% \subsection or using \label (\appendices by itself
	% starts a section numbered zero.)
	%

	% Can use something like this to put references on a page
	% by themselves when using endfloat and the captionsoff option.
	\ifCLASSOPTIONcaptionsoff
	\newpage
	\fi

	% trigger a \newpage just before the given reference
	% number - used to balance the columns on the last page
	% adjust value as needed - may need to be readjusted if
	% the document is modified later
	%\IEEEtriggeratref{8}
	% The "triggered" command can be changed if desired:
	%\IEEEtriggercmd{\enlargethispage{-5in}}
	
	% references section
	
	% can use a bibliography generated by BibTeX as a .bbl file
	% BibTeX documentation can be easily obtained at:
	% http://mirror.ctan.org/biblio/bibtex/contrib/doc/
	% The IEEEtran BibTeX style support page is at:
	% http://www.michaelshell.org/tex/ieeetran/bibtex/
	\bibliographystyle{IEEEtran}
	% argument is your BibTeX string definitions and bibliography database(s)
	\bibliography{ranking_saliency}
	
	\vspace{-1.5em}
	\begin{IEEEbiography}[{\includegraphics[width=1in,height=1.25in,clip,keepaspectratio]{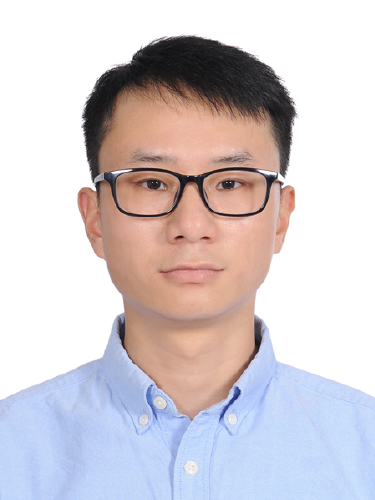}}]
		{Nian Liu} is currently a researcher with the Inception Institute of Artificial Intelligence, Abu Dhabi, UAE. He received the Ph.D. degree and the B.S. degree from the School of Automation at Northwestern Polytechnical University, in 2020 and 2012, respectively. His research interests include computer vision and machine learning, especially on saliency detection and deep learning.
	\end{IEEEbiography}

	\vspace{-1.5em}
	\begin{IEEEbiography}[{\includegraphics[width=1in,height=1.25in,clip,keepaspectratio]{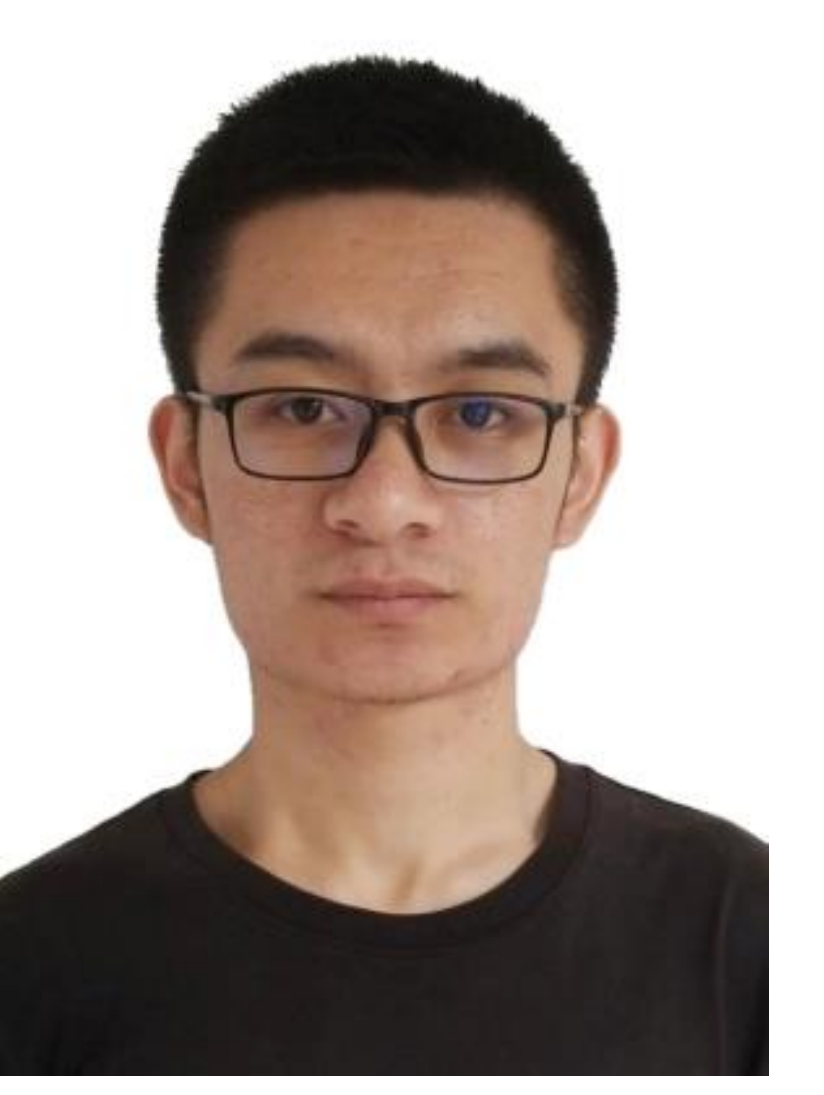}}]{Long Li} is a Ph.D. student with School of Automation at Northwestern Polytechnical University, Xi'an, China. He received the B.S. degree from Northwestern Polytechnical University in 2018. His research interests include computer vision and deep learning.
	\end{IEEEbiography}

	\vspace{-1.5em}
	\begin{IEEEbiography}[{\includegraphics[width=1in,height=1.25in,clip,keepaspectratio]{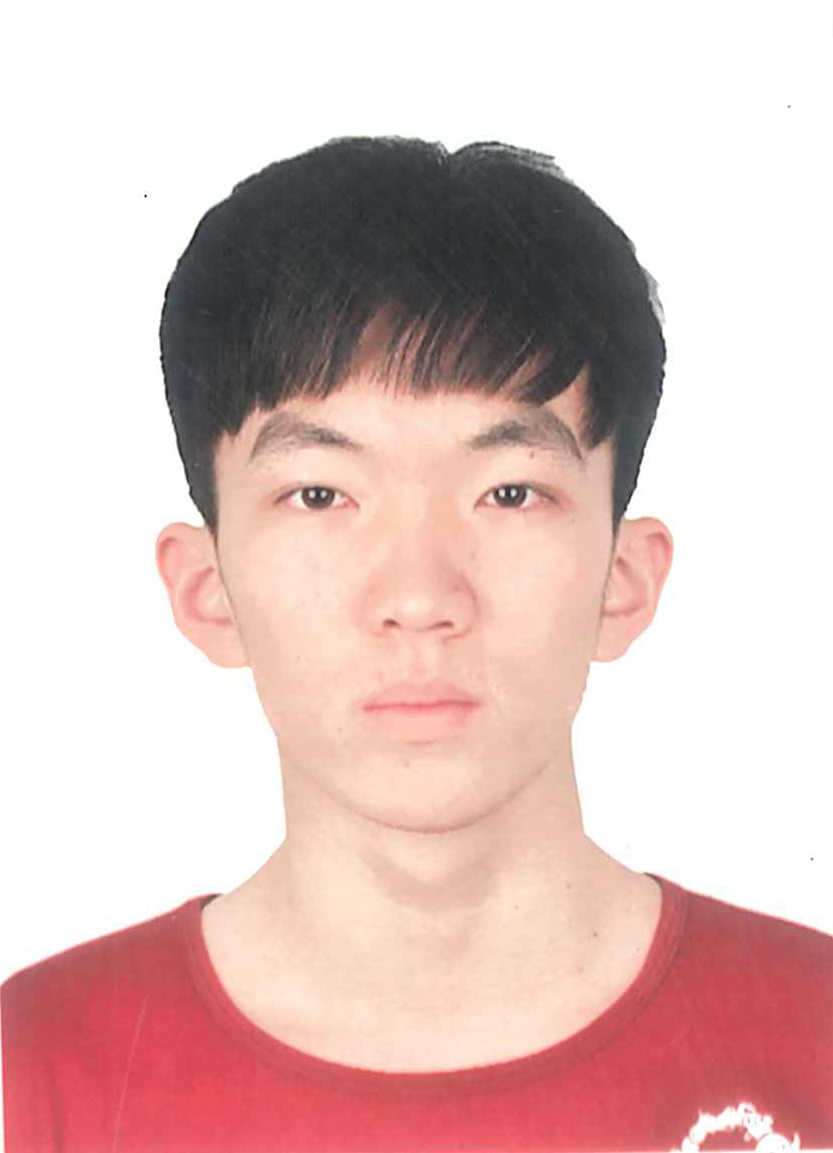}}]{Wangbo Zhao} received the B.E. degree from Northwestern Polyhtechincal University, Xi'an, China in 2019, where he is currently pursuing the master's degree. His research interest is computer vision, especially on saliency detection and video understanding.
	\end{IEEEbiography}

	\vspace{-1.5em}
	\begin{IEEEbiography}[{\includegraphics[width=1in,height=1.25in,clip,keepaspectratio]{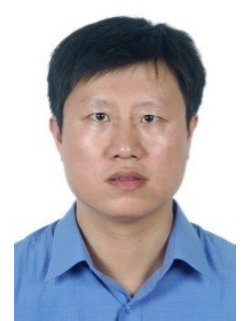}}]
		{Junwei Han} is currently a Full Professor with Northwestern Polytechnical University, Xi'an, China. His research interests include computer vision, multimedia processing, and brain imaging analysis. He is an Associate Editor of IEEE Trans. on Human-Machine Systems, Neurocomputing, Multidimensional Systems and Signal Processing, and Machine Vision and Applications.
	\end{IEEEbiography}

	\vspace{-1.5em}
	\begin{IEEEbiography}[{\includegraphics[width=1in,height=1.25in,clip,keepaspectratio]{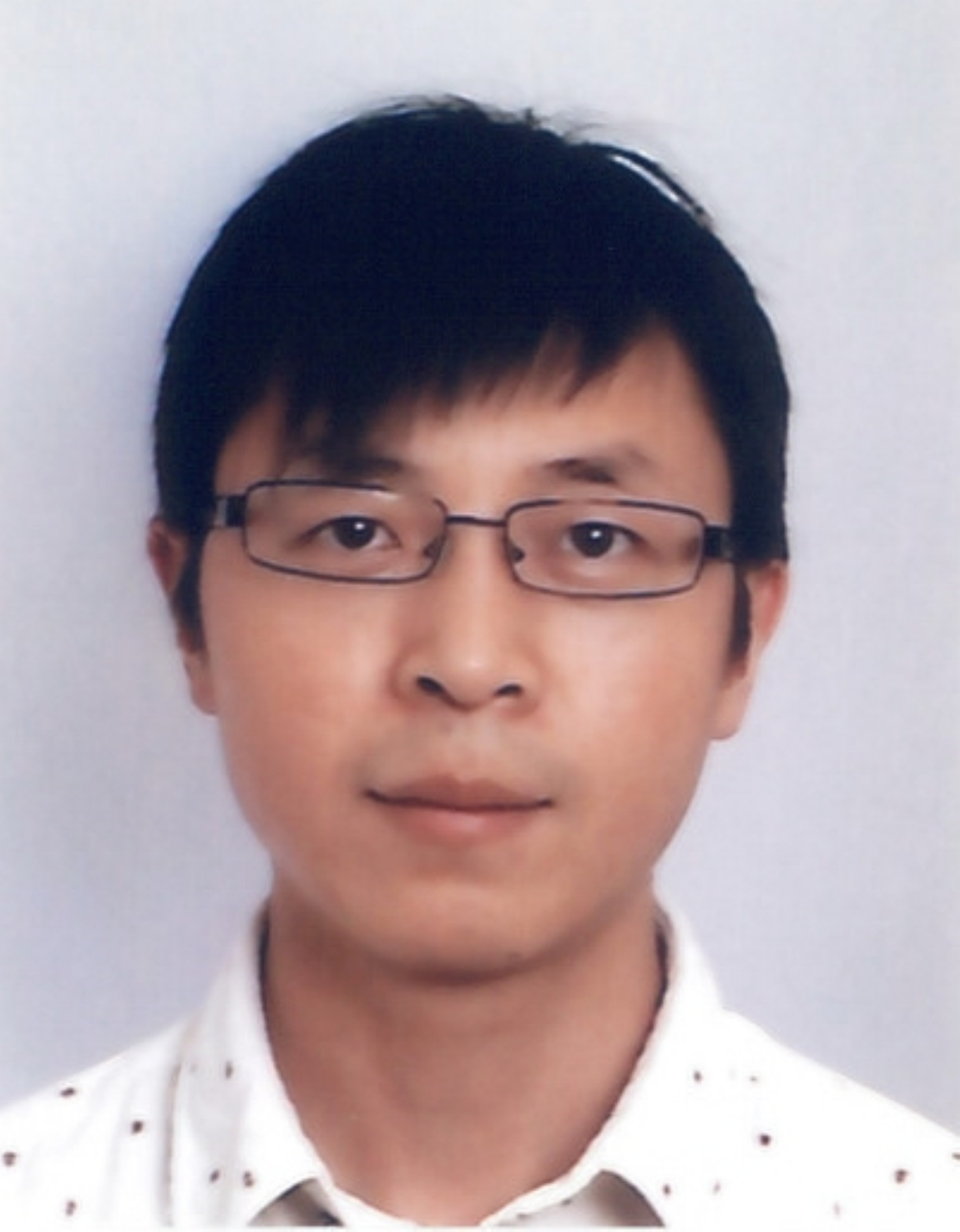}}]
		{Ling Shao} is the CEO and Chief Scientist of the Inception Institute of Artificial Intelligence (IIAI), Abu Dhabi, United Arab Emirates. His research interests include computer vision, machine learning, and medical imaging. He is a fellow of IEEE, IAPR, IET, and BCS.   
	\end{IEEEbiography}
	
	% that's all folks
\end{document}